\documentclass{ecai} 

\usepackage{latexsym}
\usepackage{amssymb}
\usepackage{amsmath}
\usepackage{amsthm}
\usepackage{booktabs}
\usepackage{enumitem}
\usepackage{graphicx}
\usepackage{color}

\usepackage{caption}
\usepackage{comment}
\usepackage{xspace}
\usepackage[most]{tcolorbox}
\tcbuselibrary{most}

\newcommand{\BibTeX}{B\kern-.05em{\sc i\kern-.025em b}\kern-.08em\TeX}

\newcommand{\propername}[1]{\mbox{\small \textsf{#1}}\xspace}
\newcommand{\ConGolog}{\propername{ConGolog}}
\newcommand{\textmath}[1]{\mbox{\textit{#1}}}
\newcommand{\Trans}{\textmath{Trans}}

\newcommand{\ndet}{\mbox{$|$}}
\newcommand{\conc}{\mbox{$\parallel$}}

\newcommand{\Math}[1]{\ensuremath{#1}}

\newcommand{\modecal}[1]{{\Math{\mathcal{#1}}}}

\newcommand{\A}{\modecal{A}} 
\newcommand{\C}{\modecal{C}} 
\newcommand{\D}{\modecal{D}}
\newcommand{\F}{\modecal{F}}
\newcommand{\N}{\modecal{N}}

\newcommand{\SitDet}{\mathit{SituationDetermined}}
\newcommand{\limp}{\supset}

\newcommand{\anyonehl}{\textsc{any1hl}}
\newcommand{\anyseqhl}{\textsc{anyseqhl}}

\newcommand*{\qedf}{\hfill\ensuremath{\blacksquare}}

\tcbset{
  myboxstyle/.style={
    colback=gray!10, 
    colframe=black, 
    fontupper=\sffamily\scriptsize\linespread{1.5}\selectfont,  
    boxrule=0.7pt, 
    enhanced, 
    title=#1, 
    coltitle=white, 
    colbacktitle=gray!90, 
    fonttitle=\bfseries, 
  }
}

\begin{document}


\begin{frontmatter}


\paperid{123} 


\title{Using Large Language Models for Abstraction of Planning
  Domains - Extended Version \\
	\LARGE{(A Preliminary Report)}}


\author[A]{\fnms{Bita}~\snm{Banihashemi}\thanks{Corresponding Author. Email: bita.banihashemi@igdore.org}}
\author[B]{\fnms{Megh}~\snm{Patel}}
\author[B]{\fnms{Yves}~\snm{Lesp\'{e}rance}} 

\address[A]{IGDORE}
\address[B]{York University}


\begin{abstract}
Generating an abstraction of a dynamic domain that aligns with a given
purpose remains a significant challenge given that the choice of such an
abstraction can impact an agent's ability to plan, reason, and provide
explanations effectively. We model the agent's concrete behaviors in
PDDL and investigate the use of in-context learning with large
language models (LLMs) for the generation of abstract PDDL domains
and problem instances, given an abstraction objective specified in
natural language. The benchmark examples we use are new
and have not been part of the data any LLMs have been trained on. We
consider three categories of abstractions: abstraction of choice of 
alternative concrete actions, abstraction of sequences of concrete
actions, and abstraction of action/predicate parameters, as well as
combinations of these. The generated
abstract PDDL domains and problem instances are then checked by
symbolic validation tools as well as human experts. Our experiments
show that GPT-4o can generally synthesize useful
planning domain abstractions in simple settings, although it is 
better at abstracting over actions than over the associated fluents.
\end{abstract}

\end{frontmatter}


\section{Introduction}

The ability to generate abstractions that ignore irrelevant details is a crucial human cognitive ability that supports reasoning and communication.
This has inspired significant research in AI, where abstraction enables systems to simplify complex tasks, enhance generalization, and improve decision-making. For instance, abstraction has been exploited to improve the efficiency of planning (e.g., \cite{DBLP:conf/aips/ChenB21}), provide explanations of agents' behavior (e.g., \cite{DBLP:conf/aips/SeegebarthMSB12}), and in reinforcement learning (e.g., \cite{DBLP:journals/ai/SuttonPS99}). 

\cite{DBLP:conf/aaai/BanihashemiGL17,DBLP:journals/ai/BanihashemiGL25} (BDL17) developed a general framework for \emph{agent abstraction} in \emph{dynamic domains} which is based on the situation calculus \cite{McCarthy1969:AI,Reiter01-Book} and the \ConGolog agent programming language \cite{DBLP:journals/ai/GiacomoLL00}. The account formalizes notions of sound/complete abstractions between a high-level (HL) action theory and a low-level (LL) action theory representing the agent's possible behaviors at different levels of detail.  These notions are based on the existence of a bisimulation relation between their respective models relative to a \emph{refinement mapping} that maps high-level fluents to low-level state formulas and high-level actions to \ConGolog programs over the low-level theory that implement them.  It is shown that sound/complete abstractions have many useful properties that allow one to reason at the high level and refine the results at the low level, and they can also be used for monitoring and explanation.
There is also much related work on hierarchical planning, for instance, hierarchical task network (HTN) planning \cite{DBLP:journals/amai/ErolHN96} and planning with domain-specific control knowledge specified in linear temporal logic (LTL) \cite{DBLP:journals/ai/BacchusK00}.

(BDL17) specifies under what conditions a high-level action theory is a sound/complete abstraction of a low-level one under a given refinement mapping,
but it does not say how one obtains such an abstraction.
In general, there would be several different abstractions of a domain, each of which serves a particular purpose and could be used in a different application. Generating an abstraction of a domain that aligns with a given purpose remains a significant challenge.
So far, there has only been limited work on solvers that can synthesize abstractions that align with a given mapping and verify their soundness and/or completeness (e.g., \cite{LuoECAI20,DBLP:conf/aaai/Luo23,DBLP:conf/aaai/FangWCLCG25,DBLP:conf/aaai/DongSZL25}).
But relying on human experts to generate such abstractions by hand is costly, time consuming, and does not scale.



%
Much recent work has investigated the use of large language models (LLMs) \cite{DBLP:journals/corr/abs-2303-18223} for reasoning about action and the generation of planning specifications. While some work shows limitations of LLMs for reasoning about actions and planning (e.g., \cite{DBLP:journals/corr/abs-2410-02162, DBLP:conf/acl/0001HX023}), other work shows good potential for the generation of Planning Domain Definition Language (PDDL) \cite{McDermot:etal:1998} specifications (e.g., \cite{DBLP:conf/icaps/OswaldSK00S24,DBLP:conf/nips/GuanVSK23}).

It is clear that LLMs embed comprehensive world knowledge and that they can generate PDDL specifications. We are interested in abstraction in reasoning about action and to see if LLMs can be used to generate useful abstractions. We are inspired by the agent abstraction framework of (BDL17), but we concentrate on abstraction generation using PDDL specifications as LLMs are commonly trained on planning domain and problem representations expressed in this formalism. Thus it seems reasonable that they could be very helpful in generating such abstractions. 



In this paper, we investigate we investigate the ability of GPT-4o \cite{DBLP:journals/corr/abs-2410-21276} to generate such abstractions. 
We define the \emph{Planning Domain Abstraction Generation} (PDAG) task as follows:
given (1) a concrete planning domain in PDDL (2) a concrete planning problem instance in PDDL, (3) a brief natural language description of the domain, and (4) a purpose of abstraction in natural language, use in-context learning with LLMs to generate
(a) an abstract planning domain in PDDL and (b) an abstract planning problem instance in PDDL 
that meet the purpose of abstraction.

We consider three basic categories of abstractions: (1) abstraction of alternative concrete actions, e.g. $book\_hotel$ and $book\_airbnb$ could be abstracted to $book\_accommodation$, (2) abstraction of sequences of concrete actions, e.g., the sequence $enter\_username$ and $enter\_password$ could be abstracted to $login$, and (3) abstraction of action/predicate parameters, e.g., abstracting over the $room\_view$ parameter of $book\_room$ and dropping associated fluents, types, and actions.
We also examine a fourth category which is a combination of (1) and (2), i.e., abstraction of alternative sequences of concrete actions.
We have developed a number of benchmark examples for the PDAG task for each of these categories.

We ran experiments to evaluate the capability of GPT-4o to solve the PDAG task for these benchmarks, which are new and have not been seen by the LLM before.
We use zero-shot and one-shot prompting \cite{DBLP:journals/corr/abs-2005-14165,DBLP:journals/corr/abs-2301-00234,DBLP:conf/ranlp/Li23}, augmented with chain-of-thought reasoning \cite{DBLP:conf/nips/Wei0SBIXCLZ22} and role-play \cite{DBLP:conf/naacl/KongZCLQSZWD24}. 
%
To evaluate the generated abstract planning domains and problems, we use the plan validator tool VAL \cite{DBLP:conf/ictai/HoweyLF04}, in addition to evaluation by human experts. We also use the Fast Downward planner \cite{DBLP:journals/jair/Helmert06} to generate plans and detect errors.

The PDAG task and our benchmarks for it are inspired by the (BDL17) framework, but we do not require the LLM to generate the refinement mapping.
%
Nor do we formally verify that the abstract PDDL domain and problem returned by the LLM is a solution to the PDAG task.
We leave these topics for future work.
But in Section \ref{sec:pdag}, we propose a formalization of the PDAG task.  This could be used to formally specify instances of the task and verify that an answer is a solution.  One could also develop a system to perform this verification automatically.

The results of our experiments show that GPT-4o can generate almost error-free results for abstraction of action/predicate parameters, 
but that it makes more mistakes as the difficulty of abstraction increases with abstraction of alternative concrete
actions, abstraction of sequences of concrete actions, and abstraction of alternative sequences of concrete actions.

The paper is organized as follows.
In the next section, we present some technical background on
the situation calculus and \ConGolog, the agent abstraction framework of (BDL17), classical planning and PDDL, and in-context learning with LLMs.
Most of this section may be skipped on a first reading 
as the material in Sections \ref{sec:AbsCategories} and \ref{sec:Implement} should be sufficient to understand how we use LLMs for the generation of abstractions.
Section \ref{sec:pdag} presents the planning domain abstraction generation task and discusses a possible formalization. Then in Section \ref{sec:AbsCategories} we discuss different categories of abstraction. Implementation details, including design of prompts and the system architecture are provided in Section \ref{sec:Implement}. In Section \ref{sec:Eval} we discuss the results of experimental evaluations. Related work is discussed in Section \ref{sec:RelatedWork}. Finally, Section \ref{sec:ConcFuture} concludes with avenues for future research. Appendices \ref{subSec:AppExamples} and \ref{subSec:Prompts} include the complete listings of examples used and sample prompts respectively.

\section{Background} \label{sec:background} 

\subsection{Situation Calculus and High-Level Programs}
\paragraph{Situation Calculus}

The \textit{situation calculus} \cite{McCarthy1969:AI,Reiter01-Book} is a well known predicate logic language
for representing and reasoning about dynamically changing
worlds \cite{McCarthy1969:AI,Reiter01-Book}. 
All changes to the world are the result of \textit{actions},
which are terms in the language. 
%
A possible world history is represented by a term called a
\emph{situation}. The constant $S_0$ is used to denote the initial
situation.
Sequences of
actions are built using the function symbol $do$, such that $do(a,s)$
denotes the successor situation resulting from performing action $a$
in situation $s$.
%
Predicates and functions whose value varies from situation to situation are
called \textit{fluents}, and are denoted by symbols taking a
situation term as their last argument.
For example, we may have that $Door1$ is not open in the initial
situation $S_0$, i.e., $\lnot IsOpen(Door1,S_0)$ holds, but is open in the
situation that results from doing the action $open(Door1)$ in $S_0$,
i.e., $IsOpen(Door1,do(open(Door1),S_0)$ holds.
$s \sqsubset s'$ means that $s$ is a  predecessor situation of $s'$, and
$s \sqsubseteq s'$ stands for $s = s' \lor s\sqsubset s'$.
The abbreviation  $do([a_1, \ldots, a_n],s)$ stands for
$do(a_n,do(a_{n-1},\ldots , do(a_1,s) \ldots ))$; also
for an action sequence $\vec{a}$, we often write $do(\vec{a}, s)$ for $do([\vec{a}],s)$.
%


Within this language, one can formulate action theories that describe
how the world changes as a result of the available actions.  Here,
we concentrate on \emph{basic action theories} 
as proposed in \cite{Reiter01-Book}.

%
%

\paragraph{ConGolog.}
To represent and reason about complex actions or processes obtained by suitably
executing atomic actions, various so-called \emph{high-level programming
languages} have been defined.
Here we concentrate on (a variant of) \ConGolog that includes the following constructs: 
\vspace{-1em}

\begin{small}
 \[ \delta ::= nil \mid \alpha  \mid  \varphi?  \mid  \delta_1;\delta_2  \mid
  \delta_1 \ndet \delta_2  \mid  \pi x.\delta  \mid  \delta^*  \mid 
  \delta_1 \conc \delta_2 \]
\end{small}%

In the above, $\alpha$ is an action term, possibly with parameters,
and  $\varphi$ is a 
situation-suppressed formula, i.e., a formula 
with all situation arguments in fluents suppressed. 
As usual, we denote by $\varphi[s]$ the 
formula obtained from $\varphi$ by restoring the situation
argument $s$ into all fluents in $\varphi$.
The sequence of program $\delta_1$ followed by program $\delta_2$ is denoted by $\delta_1;\delta_2$.
Program $\delta_1 \ndet \delta_2$ allows for the nondeterministic choice
between programs $\delta_1$ and $\delta_2$, while $\pi x.\delta$ executes program
$\delta$ for \textit{some} nondeterministic choice of a  
binding for object variable $x$ (observe that such a choice is, in general, unbounded).  $\delta^*$
performs $\delta$ zero or more times.
Program $\delta_1 \conc \delta_2$ expresses the concurrent execution
(interpreted as interleaving) of programs $\delta_1$ and $\delta_2$.
%
The construct \textbf{if} $\phi$ \textbf{then} $\delta_1$
\textbf{else} $\delta_2$ \textbf{endIf} is defined as $[\phi?;
\delta_1 ] \mid [\neg \phi?; \delta_2]$. 
We also use $nil$, an abbreviation for $True?$, to represents the
\emph{empty program}, i.e., when nothing remains to be performed. 

Formally, the semantics of \ConGolog\ is specified in terms of
single-step transitions, using the following two predicates
\cite{DBLP:journals/ai/GiacomoLL00}: 
\emph{(i)} $Trans(\delta,s,\delta',s')$, which holds if one step
of program $\delta$ in situation $s$ may lead to situation $s'$ with
$\delta'$ remaining to be executed; and \emph{(ii)}  $Final(\delta,s)$, which
holds if program $\delta$ may legally terminate in situation $s$.
The definitions of $Trans$ and $Final$ we use are as in
\cite{DBLP:journals/ai/GiacomoLL00}, except that
the test construct $\varphi?$ does not yield
any transition, but is final when satisfied.

Predicate $Do(\delta,s,s')$ means that program $\delta$, when executed starting in situation $s$, has as a legal terminating situation $s'$, and is defined as $Do(\delta,s,s') \doteq \exists \delta'. Trans^*(\delta,s,\delta',s') \land Final(\delta',s')$ where $\Trans^*$ denotes the reflexive transitive closure of $\Trans$.


\paragraph{Situation-Determined Programs.}

A \ConGolog program $\delta$ is \emph{situation-determined} (SD) in a situation $s$ \cite{DBLP:conf/aamas/GiacomoLM12}  if for every sequence of actions, the remaining program is  determined by the resulting situation, i.e.,

\noindent
\begin{small}
$\SitDet(\delta,s) \doteq \forall s',\delta', \delta''. \\$
$\hspace*{3em}\Trans^*(\delta,s, \delta',s') \land \Trans^*(\delta,s, \delta'',s') \limp$
$\delta'=\delta''$
\end{small}%

\subsection{Abstracting Agent Behavior}
In the agent abstraction framework of (BDL17),
there is a high-level (abstract) (HL) basic action
theory $\D_h$ and a low-level (concrete) (LL) basic action theory 
$\D_l$ representing the agent's possible behaviors at different levels
of detail.
$\mathit{\D_h}$ (resp. $\mathit{\D_l}$) involves a finite set of primitive action types
$\mathit{\A_h}$ (resp. $\mathit{\A_l}$) and a finite set of primitive
fluent predicates $\mathit{\F_h}$ (resp. $\mathit{\F_l}$).
The terms of object sort are assumed be a
countably infinite set $\N$ of standard names for which we have the unique
name assumption and domain closure.\footnote{This makes it easier to
  relate the high-level and low-level action theories.
One of the main consequences of assuming standard names is that quantification
can be considered substitutionally; for instance, $\exists x.P(x)$ is
true just in case $P(n)$ is true for some standard name $n$.} 
Also, $\mathit{\D_h}$ and $\mathit{\D_l}$ are assumed to share no
domain specific symbols except for the set of standard names for
objects $\mathit{\N}$.
For simplicity, and w.l.o.g., (BDL17) assume that there are no functions
other than constants and no non-fluent predicates.

\paragraph{Refinement Mapping.}
To relate the two theories, a \emph{refinement mapping} $m$ is defined as a function 
that associates each high-level primitive action type $\mathit{A}$ in $\mathit{\A_h}$ to
a 
SD \ConGolog program $\delta_A$ defined over the
low-level theory 
that implements the action, i.e., $m(A(\vec{x})) = \delta_A(\vec{x})$;
moreover, $m$ maps each situation-suppressed high-level fluent
$F(\vec{x})$ in $\mathit{\F_h}$ to a situation-suppressed formula
$\phi_F(\vec{x})$ defined over the low-level theory that characterizes
the concrete conditions under which $F(\vec{x})$ holds in a situation. 
A mapping can be extended to a sequence of actions in the
obvious way, i.e.,  $m_a(\alpha_1, \ldots, \alpha_n) \doteq m_a(\alpha_1);\ldots;
m_a(\alpha_n)$ for $n \geq 1$ and  $m_a(\epsilon) \doteq nil$. The notation can also be extended so that $m_f(\phi)$ stands for the result
of substituting every fluent $\mathit{F}(\vec{x})$ in situation-suppressed formula $\phi$ by $m_f(\mathit{F}(\vec{x}))$.


\paragraph{$m$-Bisimulation.} To relate the high-level and low-level
models/theories, a variant of bisimulation
\cite{DBLP:conf/ijcai/Milner71,DBLP:books/daglib/0067019} is defined as follows. 
%
Given $M_h$ a model of $\D_h$, and $M_l$ a model of $\D_l \cup \C$,
a relation $B \subseteq \Delta_S^{M_h} \times \Delta_S^{M_l}$ (where $\Delta_S^{M}$ stands for the situation domain of $M$)  is an
\emph{$m$-bisimulation relation between $M_h$ and $M_l$}  if $\langle s_h, s_l \rangle \in B$ implies that: $(i)$ $s_h \sim_m^{M_h,M_l} s_l$, i.e., $s_h$ and $s_l$ evaluate each high-level primitive fluent the same; $(ii)$  for every high-level primitive action type $\mathit{A}$ in $\mathit{\A_h}$, if  there exists $s_h'$ such that $M_h \models  Poss(A(\vec{x}),s_h) \land s_h' = do(A(\vec{x}),s_h)$, 
then there exists  $s_l'$ such that $M_l \models
Do(m(A(\vec{x})),s_l,s_l')$ and $\langle s_h',s_l' \rangle \in B$; and
 $(iii)$ for every high-level primitive action type $\mathit{A}$ in  $\mathit{\A_h}$, if there exists  $s_l'$ such that 
  $M_l \models Do(m(A(\vec{x})),s_l,s_l')$, then there exists $s_h'$ such that
  $M_h \models Poss(A(\vec{x}),s_h) \land s_h' =   do(A(\vec{x}),s_h)$ and $\langle s_h',s_l' \rangle \in B$. 
$M_h$ is \emph{$m$-bisimilar} to $M_l$, written $M_h \sim_m M_l$, if and only if there exists an
$m$-bisimulation relation $B$ between $M_h$ and $M_l$ such that $(S_0^{M_h}, S_0^{M_l}) \in B$.

\paragraph{Sound/complete abstractions.} In (BDL17), $\D_h$ is a \emph{sound abstraction of} $\D_l$ \emph{relative to refinement mapping} $m$ if and only if, for all models $M_l$ of $\D_l \cup \C$, there exists a model $M_h$ of $\D_h$ such that  $M_h \sim_m M_l$.
With a sound abstraction, whenever the high-level theory \emph{entails} that a sequence of actions is executable and achieves a  certain condition, then  the low level must also entail that there exists an executable refinement of the sequence such that the ``translated'' condition holds afterwards. Moreover, whenever the low level considers the executability of a refinement of a high-level action is satisfiable, then the high level does also. 
A characterization that provides the basis for
automatically verifying that one has a sound abstraction is also given.

A dual notion is also defined: $\D_h$ is a \emph{complete abstraction of} $\D_l$ \emph{relative to refinement mapping} $m$ if
and only if, for all models $M_h$ of $\D_h$, there exists a model $M_l$ of $\D_l \cup \C$ such that  $M_l \sim_m M_h$.

Now, let us define some low-level programs that characterize
the refinements of high-level action/action sequences:

\begin{small}
\[\begin{array}{l}
\mathit{\anyonehl} \doteq \ \ \mid_{A_i \in \A_h} \pi
    \vec{x}.m(A_i(\vec{x}))
    \\ 
    \qquad ~\mbox{i.e., do any refinement of any one HL primitive action,}
\\[1ex]
\mathit{\anyseqhl} \doteq \mathit{\anyonehl}^*
\\ 
\qquad ~~\mbox{i.e., do any sequence of refinements of HL actions.}
\end{array}\]
\end{small}%

How does one verify that one has a sound abstraction?  The following
result identifies necessary and sufficient conditions for having a sound
abstraction:
\\[1ex]
\textbf{Theorem} [Theorem 10 in \cite{DBLP:journals/ai/BanihashemiGL25}] 
 $\D^h$ is a sound abstraction of $\D^l$ relative to mapping $m$ if and only if
\begin{small}
\begin{description}
\item[(a)]
$\D^l_{S_0} \cup \D^l_{ca} \cup \D^l_{coa} \models m(\phi)$, for all
$\phi \in D^h_{S_0}$,
\item[(b)]
$\mathit{\D^l \cup \C \models \forall s. Do(\anyseqhl,S_0,s) \limp}$ \\
\hspace*{1em} $\bigwedge_{A_i \in \A^h}
    \forall \vec{x}.
 (m(\phi^{Poss}_{A_i}(\vec{x}))[s] 
    \equiv \exists s' Do(m(A_i(\vec{x})),s,s')),$
\item[(c)]
$\D^l \cup \C \models \forall s. Do(\anyseqhl,S_0,s) \limp $ \\
\hspace*{2em} $\bigwedge_{A_i \in \A^h}  \forall \vec{x}, s'. (Do(m(A_i(\vec{x})),s,s') \limp$ \\
\hspace*{4.1em} $\bigwedge_{F_i \in \F^h}  \forall \vec{y}  (m(\phi^{ssa}_{F_i,A_i}(\vec{y},\vec{x}))[s] \equiv m(F_i(\vec{y}))[s'])),$
\end{description}
\end{small}%
\noindent
where $\phi^{Poss}_{A_i}(\vec{x})$ is the right hand side (RHS) of the precondition axiom for action $A_i(\vec{x})$, and $\phi^{ssa}_{F_i,A_i}(\vec{y},\vec{x})$ is the RHS of the successor state axiom for $F_i$ instantiated with action
$A_i(\vec{x})$ where action terms have been eliminated using
$\D^h_{ca}$. 

\noindent
The above 
provides us with a way of showing
that we have a sound abstraction by proving that certain properties are entailed by the low-level theory.
The theorem also means that if $\D^h$ is a sound abstraction of $\D^l$ with respect to $m$, 
then $\D^l$ must entail the mapped high-level successor state axioms
and entail that the mapped conditions for a high-level action to be executable
(from the precondition axioms of $\D^h$) correctly capture the
executability conditions of their refinements (these conditions must
hold after any sequence of refinements of high-level actions, i.e.,
in any situation $s$ where $Do(\anyseqhl,S_0,s)$ holds).

  For the special case where $\D^h_{S_0}$ is a complete theory, we also have
  the following result:
  \\[1ex]
  \textbf{Corollary} [Corollary 16 in \cite{DBLP:journals/ai/BanihashemiGL25}]
  If $\D^h_{S_0}$ is a complete theory
  (i.e., for any situation suppressed formula $\phi$,
  either $\D^h_{S_0} \models   \phi[S_0]$ or
   $\D^h_{S_0} \models   \lnot \phi[S_0]$)
   and  $\D^l$ is satisfiable,
   then if $\D^h$ is a sound abstraction of $\D^l$ with respect to $m$,
   then $\D^h$ is also a complete abstraction of $\D^l$ with respect
   to $m$.

\subsection{Classical Planning and PDDL.}
Classical Planning \cite{DBLP:books/daglib/0014222} is the task of determining a sequence of actions that transforms a given initial state to a state satisfying a goal condition  
within a discrete, deterministic, and fully observable environment. 

The Planning Domain Definition Language (PDDL) \cite{McDermot:etal:1998} is a widely used language for representing classical planning problems. In this paper, we use the STRIPS fragment of PDDL with typing. PDDL separates the definition of a planning problem into two parts: the \emph{domain} definition and a \emph{problem} definition. 

The \emph{domain} definition provides a lifted representation of the relevant aspects and underlying rules of the world. It includes a hierarchy of \emph{types}, in addition to \emph{predicates} and \emph{actions}. 
%
\emph{Types} control how parameters may be instantiated by object names in the problem definition. 
\emph{Predicates} define the state space of the model, and contain the list of the model's boolean variables (i.e., facts that are either true or false).
\emph{Actions} trigger transitions between states, and typically include preconditions and effects. 
Preconditions define when the action is applicable, and 
in the STRIPS fragment of PDDL  
they are represented by conjunction of atoms.  
Effects define how the state changes if the action were applied, and they are represented by conjunction of positive and negative literals. 


The \emph{problem} definition models a particular problem instance in the domain which specifies the initial state, goal condition, and the object names. 
\emph{Object names} constitute the values that parameters in the domain definition may take (i.e., that ground the domain). 
The \emph{initial state specification} lists all the facts that are true in the initial state, and any fact not explicitly mentioned is assumed to be false. 
The \emph{goal} specifies a condition that has to be satisfied at the end of a
valid plan. In the STRIPS fragment of PDDL, the goal is a conjunction
of facts. A solution to such a problem is an ordered sequence
of actions that, when executed from the initial state, results in a state
that satisfies the goal condition, i.e., a valid \emph{plan} for the problem.

\subsection{In-Context Learning.}
This refers to the ability of LLMs to generalize to novel 
and diverse 
tasks by interpreting examples or task descriptions provided within the input prompt, without requiring explicit task-specific training or parameter fine-tuning \cite{DBLP:journals/corr/abs-2005-14165,DBLP:journals/corr/abs-2301-00234,DBLP:conf/ranlp/Li23}. In \emph{one-shot} learning, the LLM is provided with a single task demonstration, along with a natural language description of the task. The example task is given together with its desired completion, and then a final task instance, i.e, user query,  is provided to the LLM, and the LLM is expected to generate the completion for it. \emph{Zero-shot} learning is similar to one-shot learning, except that no task demonstrations are provided.
In this paper, we use zero-shot and one-shot prompts that are augmented with chain-of-thought (CoT) \cite{DBLP:conf/nips/Wei0SBIXCLZ22} and role-play \cite{DBLP:conf/naacl/KongZCLQSZWD24} prompting techniques. CoT facilitates step-by-step reasoning by encouraging the model to generate intermediate reasoning steps before arriving at a final answer, thereby improving performance on commonsense and symbolic reasoning tasks. Role-Play prompting enhances task understanding by guiding the model to assume specific personas  
or expert perspectives, 
which helps contextualize responses more effectively.

\section{Planning Domain Abstraction Generation} \label{sec:pdag}

We define the \emph{Planning Domain Abstraction Generation} (PDAG) task as follows:
given (1) a concrete planning domain in PDDL $D_c$, (2) a concrete planning problem instance in PDDL $P_c$, (3) a brief natural language description of the domain $dd_c$, and (4) a purpose of abstraction in natural language $pa$,
use in-context learning with LLMs to generate
(a) an abstract planning domain in PDDL $D_a$ and (b) an abstract planning problem instance in PDDL $P_a$
that meet the purpose of abstraction.

We can formalize the PDAG task based on the notion of sound and complete abstraction of situation calculus basic action theories (BATs) relative to a refinement mapping $m$ of (BDL17). 
For this, the purpose of abstraction must be expressed as a set of constraints $C_m$ on the refinement mapping $m$ (these constraints restrict which fluents and actions may be included in the abstract theory and how they can be mapped to the concrete theory; see the next section for an example).
First, note that for any domain $D$ and problem $P$ in the ADL \cite{DBLP:conf/kr/Pednault89} fragment of PDDL, we can obtain an equivalent BAT $t(D,P) = \mathcal{D}$ as shown in \cite{DBLP:conf/ijcai/ClassenELN07}.  Note that this BAT must have a finite object domain and that the initial situation description must be complete, i.e., it completely specifies which tuples of objects are in the extension of every fluent initially.
Using this, we say that an abstract PDDL domain $D_a$ and problem $P_a$ is a \emph{solution to the PDAG task} $(D_c, P_c, C_m)$, where we have the concrete PDDL domain $D_c$ and problem $P_c$ and constraints on the mapping $C_m$, if there exists a refinement mapping $m$ such that $m$ satisfies $C_m$ and $t(D_a,P_a)$ is a sound and complete abstraction of $t(D_c,P_c)$ relative to $m$.  \cite{DBLP:journals/ai/BanihashemiGL25} identifies a set of necessary and sufficient properties for an abstract  BAT $\mathcal{D}_a$ to be a sound and complete abstraction of a concrete BAT $\mathcal{D}_c$ relative to mapping $m$.  Since the action theories involved have a finite object domain and a complete initial state specification, given a mapping $m$, it should be possible to check if these properties are satisfied using model checking techniques.  But this requires identifying a suitable mapping $m$ and one also must show that $m$ satisfies the constraints $C_m$.
In the next section, we sketch how an instance of the PDAG task, Example 1,  could be specified formally following this approach.


%
\section{Abstraction of PDDL Domains and Benchmark
  Problems} \label{sec:AbsCategories}

In our experiments, we use a collection of new examples that we have developed and hence, have not been part of the training data of GPT-4o. 
Each example includes 4 components: a low-level PDDL domain specification, a
low-level PDDL problem instance, a brief natural language description of the domain, and a brief natural language description of purpose of abstraction.
We consider the following categories of abstraction tasks: abstraction
of alternative concrete actions, abstraction of a sequence of concrete
actions, abstraction of action/predicate parameters, and a combination of the
first two, i.e., abstraction of alternative sequences of concrete actions.

\paragraph{Abstraction of Alternative Concrete Actions.} 
In scenarios where the low-level domain offers multiple alternative
actions to achieve a subgoal, such actions can be abstracted into a
single high-level action. This is typically accompanied by the
abstraction of associated predicates and types. Note that depending on
the purpose of abstraction, the concrete domain, and the planning
goal, some types, predicates, or actions may not need to be abstracted.

\paragraph{Example 1.} \label{ex:AbsAlterLL}
Consider the following low-level domain which models travel arrangements.
In order to go to the destination, one may book a flight or train, if a seat is available,
and also book a hotel or airbnb, provided that a room is
available.

\noindent
\begin{lstlisting} [numbers=none, 
                    basicstyle=\scriptsize\sffamily, 
										xleftmargin=0.2em, 
										columns=fixed,       
                    keepspaces=true,
										breaklines=true,
										postbreak=\mbox{\textcolor{blue}{$\hookrightarrow$}\space\space},										
										label={lst:Cat2TravelLL}]
(define (domain travelArrange01_LL)
(:requirements :strips :typing)
(:types 
    hotel airbnb room flight trainRide seat - object)
(:predicates
    (booked_hotel  ?r -room  ?h - hotel)
    (booked_airbnb  ?r - room ?ab - airbnb)
    (available_room_hotel  ?r  - room ?h - hotel)
    (available_room_airbnb  ?r - room ?ab - airbnb)
    (bookedHotelOrAirbnb)
    (available_seat_flight ?s - seat ?f - flight)
    (available_seat_trainRide ?s -seat ?t - trainRide)
    (booked_flight ?s - seat ?f - flight)
    (booked_trainRide ?s - seat ?t - trainRide)
    (bookedFlightOrTrainRide))
(:action book_hotel
    :parameters (?h - hotel ?r - room)
    :precondition (available_room_hotel ?r ?h)
    :effect (and (booked_hotel ?r ?h) 
                 (not (available_room_hotel ?r ?h))
                 (bookedHotelOrAirbnb)))    
(:action book_airbnb
    :parameters (?ab - airbnb ?r - room)
    :precondition (available_room_airbnb ?r ?ab)                          
    :effect (and (booked_airbnb ?r ?ab) 
                 (not (available_room_airbnb ?r ?ab))
                 (bookedHotelOrAirbnb)))      
(:action book_flight
    :parameters (?f - flight ?s - seat)
    :precondition (available_seat_flight ?s ?f)                          
    :effect (and (booked_flight ?s ?f) 
                 (not (available_seat_flight ?s ?f))
                 (bookedFlightOrTrainRide)))     
(:action book_trainRide
    :parameters (?t - trainRide ?s - seat)
    :precondition (available_seat_trainRide ?s ?t)                          
    :effect (and (booked_trainRide ?s ?t) 
                 (not (available_seat_trainRide ?s ?t))
                 (bookedFlightOrTrainRide))))
\end{lstlisting}

\noindent
The goal in the associated problem instance is defined as follows (see Appendix \ref{subSec:AppExamples} for the complete listings
of concrete and abstract domains and problem instances of examples 1 to 4)): 

\begin{lstlisting} [numbers=none, 
                    basicstyle=\scriptsize\sffamily, 
										xleftmargin=0.2em, 
										columns=fixed,       
                    keepspaces=true,
										breaklines=true,
										postbreak=\mbox{\textcolor{blue}{$\hookrightarrow$}\space\space},										
										label={lst:GoalTravel}]										
(:goal (and (bookedFlightOrTrainRide) 
            (bookedHotelOrAirbnb))
\end{lstlisting}%

The purpose of abstraction may be expressed in natural language as
``provide an abstraction of the concrete domain such that the
high-level domain abstracts over booking various kinds of
transportation and accommodation''.

In such a case, we would expect the LLM to generate the following high-level domain:

\noindent
\begin{lstlisting} [numbers=none, 
                    basicstyle=\scriptsize\sffamily, 
										xleftmargin=0.2em, 
										columns=fixed,       
                    keepspaces=true,
										breaklines=true,
										postbreak=\mbox{\textcolor{blue}{$\hookrightarrow$}\space\space},										
										label={lst:Cat2TravelHL}]
(define (domain travelArrange01_HL)
(:requirements :strips :typing)
(:types accommodation room transportation seat 
 - object)
(:predicates
    (booked_accommodation  ?r - room ?a - accommodation)
    (available_room  ?r - room ?a - accommodation)
    (doneBookingAccommodation)
    (available_seat ?s - seat ?tp - transportation)
    (booked_transportation ?s - seat ?tp - transportation)
    (doneBookingTransportation))
(:action book_accommodation
    :parameters (?a - accommodation ?r - room)
    :precondition (available_room ?r ?a)                          
    :effect (and (booked_accommodation ?r ?a) 
                 (not (available_room ?r ?a))
                 (doneBookingAccommodation)))      
(:action book_transportation
   :parameters (?tp - transportation ?s - seat)
   :precondition (available_seat ?s ?tp)                          
   :effect (and (booked_transportation ?s ?tp) 
                (not (available_seat ?s ?tp))
                (doneBookingTransportation))))
\end{lstlisting}

\noindent
In the above, the types $hotel$ and $airbnb$ have been
abstracted to type $accommodation$, actions $book\_hotel$ and
$book\_airbnb$ to abstract action $book\_accommodation$, and
predicates $booked\_hotel$ and $booked\_airbnb$ to high-level
predicate $booked\_accommodation$. The type $room$ on the other hand,
has not been abstracted.

The problem instance can then be abstracted by using the types,
predicates, and actions in the high-level domain. Objects also need to
be assigned to the abstract types when necessary. 

In the agent abstraction framework of (BDL17) the purpose of abstraction might be specified formally as a set of
constraints on the mapping $m$, which should be
entailed by the concrete action theory $\mathcal{D}_c$ obtained from
the concrete PDDL domain and problem above.
For the booking of accommodation, we could have:
\vspace{-1.1em}

\begin{small}
\[\begin{array}{l}
    \exists r,h.Booked\_hotel(r,h,s) \lor \exists r,ab.Booked\_airbnb(r,ab,s)\\
    \quad \supset m(DoneBookingAccomodation)[s]\\
    Do(m(book\_accomodation(a,r)),s,s') \supset \\
    \quad m(DoneBookingAccomodation)[s']
  \end{array}\]
\end{small}

\noindent
This says that booking a hotel or an airbnb is sufficient to achieve
the concrete condition into which $DoneBookingAccomodation$ is mapped and
that executing the program into which the abstract action
$book\_accomodation(a,r)$ is mapped achieves $DoneBookingAccomodation$
($Do(\delta,s,s')$ means that there is an execution
of the program $\delta$ starting in situation $s$ and ending in situation $s'$).
We could also have a constraint that ensures that there is no abstract action
$a_h$ that is mapped into the concrete action $book\_hotel(h,r)$,
i.e. $m(a_h(h,r)) = book\_hotel(h,r)$, and thus this action cannot
appear at the abstract level, and similarly for $book\_airbnb$.
Also, additional constraints may require that there are no abstract predicates
that map to $Booked\_hotel$, $Booked\_airbnb$, $Hotel$, and $Airbnb$. 
The constraints for the booking of transportation would be similar.
$\qedf$

This category includes eight benchmark examples, with domains modeling diverse scenarios such as apparel selection, daily activities, household chores, and software development. The number of actions defined within these domains ranges from 3 to 6. 
%


\paragraph{Abstraction of Sequences of Concrete Actions.} 
In this category, sequences of two or more low-level actions must be
abstracted to one high-level action. This typically also involves
abstraction of conjunction of several low-level predicates into a
single high-level predicate. Similarly to the previous category, 
some predicates or actions need not be abstracted.


\paragraph{Example 2.} \label{ex:AbsSeqLL} 
Consider the following concrete domain which models editing a file in an online editor hosted in a cloud.
A user needs to first enter a valid username and then a valid matching password to login to his cloud account.
Then he can open the file in an editor provided the file is originally closed and he has permission to edit it and after that, make changes to the file content (see \ref{subsubSec:AbsSeqLLProblem} for complete listing).

\noindent
\begin{lstlisting} [numbers=none, 
                    basicstyle=\scriptsize\sffamily, 
										xleftmargin=0.2em, 
										columns=fixed,       
                    keepspaces=true,
										breaklines=true,
										postbreak=\mbox{\textcolor{blue}{$\hookrightarrow$}\space\space},
										label={lst:Cat3CloudLL}]
(define (domain cloudApps01_LL)
...
 (:action enter_UserName
     :parameters (?u - userName)
     :precondition (valid_userName ?u)
     :effect (authenticated_userName ?u))
 (:action enter_passWord
     :parameters (?u - userName ?p - passWord)
     :precondition (and (valid_passWord ?p)
                        (authenticUserPassword ?u ?p)                     
                        (authenticated_userName ?u))                          
     :effect (authenticated_passWord ?p))      
 (:action openFileInEditor
     :parameters (?f - file ?p - passWord ?u - userName)
     :precondition (and (closed_file ?f)
                        (hasEditPermission ?u ?f) 
                        (authenticated_passWord ?p))                          
     :effect (and (openedFileInEditor ?f)
                  (not(closed_file ?f)) ))   
 (:action changeFileContent
     :parameters (?f - file)
     :precondition (openedFileInEditor ?f)                          
     :effect (changedFileContent ?f)))     
\end{lstlisting}%

Now suppose that the purpose of abstraction states ``provide an
abstraction of the concrete domain such that the high-level domain
abstracts over detailed steps of logging in and editing a file on
cloud''. The following high-level domain satisfies the purpose of
abstraction:

\begin{lstlisting} [numbers=none, 
                    basicstyle=\scriptsize\sffamily, 
										xleftmargin=0.2em, 
										columns=fixed,       
                    keepspaces=true,
										breaklines=true,
										postbreak=\mbox{\textcolor{blue}{$\hookrightarrow$}\space\space},
										label={lst:Cat3CloudHL}]
(define (domain cloudApps01_HL)
...
(:action login
     :parameters (?u - userName ?p - passWord)
     :precondition (valid_credentials ?u ?p)
     :effect (logged_in ?u ?p))
(:action edit_file
     :parameters (?f - file ?p - passWord ?u - userName)
     :precondition (and (closed_file ?f)
                        (hasEditPermission ?u ?f)
                        (logged_in ?u ?p) )                          
      :effect (and (edited_file ?f) 
                   (not(closed_file ?f)) ) ) ) 
\end{lstlisting}

\noindent
Here, the sequence of actions $enter\_userName$ and
$enter\_passWord$ has been abstracted to $login$, and the predicate
$logged\_in$ (which abstracts over the conjunction of predicates
$authenticated\_userName$ and $authenticated\_passWord$) is a
precondition for the high-level action $edit\_file$. Note that
predicates $hasEditPermission$ and $closed\_file$ are retained in the
abstract domain.
The problem instance is then abstracted by using the types, predicates, and actions in the high-level domain as before.
$\qedf$

This category includes 11 examples, with domains modeling diverse scenarios such as car manufacturing, cooking, painting, order delivery, laptop purchase, and rescue robot. The number of actions defined within these domains ranges from 3 to 6. 

\paragraph{Abstraction of Action/Predicate Parameters.} 
This category involves modifications to the low-level PDDL domain by
abstracting away one or more action/predicate parameters.
Typically, this includes the removal of a PDDL type, which often
results in the corresponding deletion or modification of predicates
and actions that utilize this type, for instance dropping parameters
of this type.

\paragraph{Example 3.} \label{ex:AbsFeatLL}
Consider the following low-level domain which models hotel bookings.
A hotel room is characterized by room type (e.g., single or double), and a room view (e.g., ocean view or garden view). It is possible to modify a room's type, for instance, convert a single room to a double room. 

\noindent
\begin{lstlisting} [numbers=none, 
                    basicstyle=\scriptsize\sffamily, 
										xleftmargin=0.2em, 
										columns=fixed,       
                    keepspaces=true,
										breaklines=true,
										postbreak=\mbox{\textcolor{blue}{$\hookrightarrow$}\space\space},
										label={lst:Cat1HBLL}]
(define (domain travelArrange02_LL) 
...
(:action book_hotel 
  :parameters (?h - hotel ?r - room ?rv - r_view ?rt - r_type) 
  :precondition (and (available_room_hotel ?r ?h) 
                     (roomType ?r ?rt)  
                     (roomView ?r ?rv))  
  :effect (and (booked_hotel ?rt ?h ?rv) 
               (not (available_room_hotel ?r ?h))))      
(:action  change_RoomType  
   :parameters (?r - room ?rt1 ?rt2 - r_type) 
   :precondition (and (roomType ?r ?rt1) 
                      (difRoomType ?rt1 ?rt2)                      
   :effect (and (roomType ?r ?rt2) 
                (not (roomType ?r ?rt1))))   
\end{lstlisting}%



Suppose that the purpose of abstraction is defined as ``provide an
abstraction of the concrete domain such that the high-level domain
must not include information about room view''. Then we expect the LLM
to abstract over the room view, and generate the following high-level domain, where the type $r\_view$ has been removed, and predicates that utilize $r\_view$ have been either adjusted or removed:

\noindent
\begin{lstlisting} [numbers=none, 
                    basicstyle=\scriptsize\sffamily, 
										xleftmargin=0.2em, 
										columns=fixed,       
                    keepspaces=true,
										breaklines=true,
										postbreak=\mbox{\textcolor{blue}{$\hookrightarrow$}\space\space},										
										label={lst:Cat1HBHLRV}]
(define (domain travelArrange02_HL)
...
(:action book_hotel
    :parameters (?h - hotel ?r - room ?rt - r_type)
    :precondition (and (available_room_hotel ?r ?h)
                       (roomType ?r ?rt))
    :effect (and (booked_hotel ?rt ?h) 
                 (not (available_room_hotel ?r ?h))))    
(:action change_RoomType
    :parameters (?r - room ?rt1 ?rt2 - r_type)
    :precondition (and (roomType ?r ?rt1)
                       (difRoomType ?rt1 ?rt2))                          
    :effect (and (roomType ?r ?rt2) 
                 (not (roomType ?r ?rt1))))
\end{lstlisting}


If instead we ask the LLM to abstract over the room type, then in
addition to removing $r\_type$ and adjusting/removing the associates
predicates, we also expect the LLM to eliminate the action $change\_RoomType$:

\noindent
\begin{lstlisting} [numbers=none, 
                    basicstyle=\scriptsize\sffamily, 
										xleftmargin=0.2em, 
										columns=fixed,       
                    keepspaces=true,
										breaklines=true,
										postbreak=\mbox{\textcolor{blue}{$\hookrightarrow$}\space\space},
										label={lst:Cat1HBHLRT}]
(define (domain travelArrange03_HL)
...
(:action book_hotel
    :parameters (?h - hotel ?r - room ?rv - r_view)
    :precondition (and (available_room_hotel ?r ?h)
                       (roomView ?r ?rv))    
    :effect (and (booked_hotel ?h ?rv) 
                 (not (available_room_hotel ?r ?h)))))
\end{lstlisting}

The problem instance can then adapted accordingly. 
$\qedf$

This category includes 10 examples with domains modeling diverse scenarios such as email composition, flight reservation, library, technical report writing, and travel (hotel) arrangements. The number of actions defined within these domains ranges from 1 to 5.

\paragraph{Abstraction of Alternative Sequences of Concrete Actions}
This category involves cases where the low-level domain contains
multiple action sequences that provide alternative pathways to
achieving a subgoal, and we require each such alternative sequence of actions to be abstracted into a single action.

\paragraph{Example 4.} \label{ex:AbsSeqAltLL}
Consider the following partial low-level domain specification which models alternative ways for holding a workshop at a campus: either a lecture hall is scheduled and the workshop is offered on campus, or web conferencing software is installed and the workshop is held online. 

\noindent
\begin{lstlisting} [numbers=none, 
                    basicstyle=\scriptsize\sffamily, 
										xleftmargin=0.2em, 
										columns=fixed,       
                    keepspaces=true,
										breaklines=true,
										postbreak=\mbox{\textcolor{blue}{$\hookrightarrow$}\space\space},
										label={lst:Cat4EdLLPart}]
 (define (domain education01_LL)
 ...
   (:action scheduleLectureHall
        :parameters (?w - workshop ?lh - lectureHall)
        :precondition (lecturerSelected ?w)
        :effect (lectureHallScheduled ?w ?lh) )        
   (:action lectureOnCampus
        :parameters (?w - workshop ?lh - lectureHall)
        :precondition (lectureHallScheduled ?w ?lh)
        :effect (and (lecturedOnCampus ?w ?lh)
                     (workshopOffered ?w)) ) 
   (:action installVideoConferencing
        :parameters (?w - workshop ?s - webConferenceSoftware)
        :precondition (lecturerSelected ?w)                         
        :effect (installedVideoConferencing ?w ?s) )
   (:action lectureOnline
        :parameters (?w - workshop ?s - webConferenceSoftware)
        :precondition (installedVideoConferencing ?w ?s)
        :effect (and (lecturedOnline ?w ?s)
                (workshopOffered ?w)) ) 
...
\end{lstlisting}%

Suppose that the purpose of abstraction includes the instruction to
``abstract over delivering a workshop session''.  Then the following
high-level action satisfies this purpose of abstraction:

\noindent
\begin{lstlisting} [numbers=none, 
                    basicstyle=\scriptsize\sffamily, 
										xleftmargin=0.2em, 
										columns=fixed,       
                    keepspaces=true,
										breaklines=true,
										postbreak=\mbox{\textcolor{blue}{$\hookrightarrow$}\space\space},
										label={lst:Cat4EdHLPart}]
 (define (domain education01_HL)
 ...
   (:action conductWorkShop
        :parameters (?w - workshop ?tp - teachingPlatform)
        :precondition (lecturerSelected ?w)
        :effect (and (teachingCompleted ?w ?tp)
                     (workshopOffered ?w)) ) ...	
\end{lstlisting}%

\noindent
Here, the 
type $teachingPlatform$ abstracts over the disjunction of types $lectureHall$
and $webConferenceSoftware$, and the predicate $teachingCompleted$
abstracts over the disjunction of  the conjunction of predicates
$lectureHallScheduled$ and $lecturedOnCampus$ when teaching platform
is $lectureHall$, and the conjunction of  
predicates  $installedVideoConferencing$ and $lecturedOnline$ when teaching platform refers to $webConferenceSoftware$.
The problem instance is adapted accordingly. 
$\qedf$

This category includes 6 examples with domains modeling diverse scenarios such as event planning, travel arrangements, repair robot,
beverage preparation, and choosing apparel. The number of actions defined within these domains ranges from 4 to 12.

\section{Implementation} \label{sec:Implement}

\subsection{Developing Prompts}
%
Our approach relies on one-shot and zero-shot learning, combined with CoT and role-play.
%
In the chat completions API of OpenAI's GPT-4o, prompts can be created by providing an array of messages that contain instructions for the model. It is possible to assign a different role to each message, and as a result, influence how the model might interpret the input. Here, we consider three roles: \emph{system}, \emph{user}, and \emph{assistant}.
A \emph{system prompt} sets the overarching context, behavior, or persona for the responses generated. A \emph{user prompt} contains the specific instructions or queries that a user provides to an LLM to elicit a desired response. An \emph{assistant prompt} 
represents the LLM's response to user inputs, and is conditioned on the context provided by both the system and user prompts. 
Assistant prompts also serve as demonstrations of desired outputs when using in-context learning (e.g., providing example completions in one-shot prompts).

\paragraph{Abstraction of Alternative Concrete Actions: Zero-Shot Prompt.} 
The system prompt (see figures \ref{fig:cat2SystemPromptA-P1} and \ref{fig:cat2SystemPromptA-P2} in Appendix \ref{subSec:AbsAltMapPrompts})
requires the LLM to assume the role of a PDDL expert, and further instructs the LLM to reason about the task in two stages: abstraction of domain and abstraction of problem instance. In the first stage, it instructs the LLM to consider the purpose of abstraction and reason about generating the high-level domain ontology (types, predicates and actions), by combining or generalizing related elements, with the help of  partial examples (e.g., two types 'hotel' and 'airbnb' can be combined into 'accommodation'). Note that inclusion of partial examples is necessary to guide the LLM towards the intended abstraction as our desired abstraction has not been part of any LLM's training data.
In the second stage, the LLM is instructed to use the generated high-level domain components to generate the abstract problem instance. The LLM is required to follow a number of rules which include trying to minimize the number of domain ontology elements, using terminology for abstracted domain elements that preserves focus of the domain, complying with the STRIPS fragment of PDDL, and ensuring that the goal of the problem instance remains logically consistent with the abstract domain's purpose.
%
A user's query is represented by a user prompt which includes a brief description of the low-level domain, low-level domain and problem files in PDDL, in addition to the purpose of abstraction (see Fig. \ref{fig:cat2AUserPrompt} in \ref{subSec:AbsAltMapPrompts}).
Note that the partial examples in the system prompt are based on a different topic area than the PDDL domains/instances provided in user queries.

\paragraph{Abstraction of Alternative Concrete Actions: One-Shot Prompt.}
Here, the system prompt is similar to the zero-shot prompt above, with the difference that 
 partial examples have been removed, and instead the LLM is instructed to learn from the example provided, referred to as $Case 1$ (see Fig. \ref{fig:cat2SystemPromptOneShot-P1} in \ref{subSec:AbsAltMapPrompts}). The one-shot example is modeled by a pair of user and assistant prompts 
(outlined in figures \ref{fig:cat2BUserPrompt} and \ref{fig:cat2BUserPrompt2}, and figures \ref{fig:cat2BAssistantPrompt} and \ref{fig:cat2BAssistantPrompt2} in \ref{subSec:AbsAltEapPrompts} respectively).  
%
The user prompt includes a low-level domain and problem instance in PDDL, purpose of abstraction, and a brief explanation of the domain. In addition to the high-level domain and problem instance in PDDL, the assistant prompt includes a rationale for making choices of abstraction of actions, predicates and types. 
The user's query (referred to as $Case 2$) is modelled as a user prompt similar to the zero-shot case above. Note that the topic area in $Case 1$ differs from that in $Case 2$.


\paragraph{Abstraction of Sequences of Concrete Actions.}
Here we only consider one-shot prompts as our preliminary experiments with zero-shot learning did not show promising results. 
Similar to the previous category, the system prompt instructs the LLM to follow 
a two-stage reasoning process: abstraction of domain and abstraction of problem instance. This process is guided by an example ($Case1$) and a set of predefined rules, which are mainly similar to the previous category (see figures \ref{fig:cat3SystemPrompt-P1} and \ref{fig:cat3SystemPrompt-P2} in \ref{subSec:AbsSeqPrompts}). Here, the LLM is asked to review the low-level PDDL domain and purpose of abstraction to identify important sequences of actions that should be merged into high-level
actions that achieve the same effect but eliminate unnecessary intermediate steps. 
The one-shot example and user query are modelled similarly to the case of one-shot learning in the previous category.


\paragraph{Abstraction of Action/Predicate Parameters.}
In this category, we only use a zero-shot prompting approach, as it proved sufficient for generating mostly correct results.
The system prompt (see Fig. \ref{fig:cat1SystemPrompt} in \ref{subSec:AbsFeaturesPrompts}) 
instructs the LLM to consider the purpose of abstraction and distinguish between the 
domain elements (parameters, types, predicates, actions)
which are important with respect to this purpose and must be retained in the high-level domain, and those that are not and must be eliminated. The user query for this category is defined similar to the previous zero-shot prompt (see Fig. \ref{fig:cat1UserPrompt} in \ref{subSec:AbsFeaturesPrompts}).

\paragraph{Abstraction of Alternative Sequences of Concrete Actions.} 
Here we restrict our analysis to one-shot prompts, as initial experiments with zero-shot prompts produced unsatisfactory outcomes. 
The system prompt is structured similar to previous categories' system prompts for one-shot learning, however, it 
is more generic and includes abstraction instructions which combine elements from all previous categories (see figures \ref{fig:cat4SystemPrompt1} and \ref{fig:cat4SystemPrompt2} in \ref{subSec:AbsALtSeqPrompts}). The user query and one-shot demonstration are modelled similar to previous categories.



\subsection{Evaluating the Generated Models}


We use a hybrid evaluation approach that incorporates both human evaluation and automated tools. Our method first uses VAL \cite{DBLP:conf/ictai/HoweyLF04} to 
check for syntax errors in the generated high-level domain and problem instance, and then calls the Fast Downward planner \cite{DBLP:journals/jair/Helmert06} to try to generate a plan. 


Next,  
the generated high-level domain and problem instance are reviewed by human domain experts.
Each generated high-level domain undergoes a comparative analysis with a sample abstract domain (already created by a knowledge engineer) which represents a sound abstraction of the low-level domain in user's query, and it is part of the solution to the PDAG task. 
This evaluation considers the correctness of abstract actions, predicates, types, and parameters generated, as well as the removal of actions, predicates, types, and parameters from the high-level domain that are considered unnecessary details wrt the purpose of abstraction. Concrete domain elements that must be retained in the generated abstract domain are also considered. Syntax errors are also noted. 
High-level problem instances are evaluated similarly. 
  
Human expert evaluation is essential 
as there may be more than one correct approach to abstract a domain. For example, in case of ``CookFood01''  when abstracting over sequence of low-level actions $wash, chop, marinate, grill$ using two high-level abstract actions $prepare$ and $cook$, one solution may abstract sequence of concrete actions $wash,chop$ to action $prepare$, while another solution may consider $wash,chop,marinate$ as refinement of action $prepare$. Moreover, the LLM often generates names for actions/predicates/types that may be different from those in the sample solution provided.
So this part of the evaluation would be difficult to automate.

As our aim in this paper is to evaluate the feasibility of using advanced LLMs to generate abstract planning domain and problem files based on an initial input prompt,
we do not consider providing corrective feedback from users or validation tools to the LLM to allow it to fix errors and re-generate high-level domains and/or problem files.

\subsection{System Architecture}

\begin{figure}[htp]
	\centering
		\includegraphics[scale=0.40]{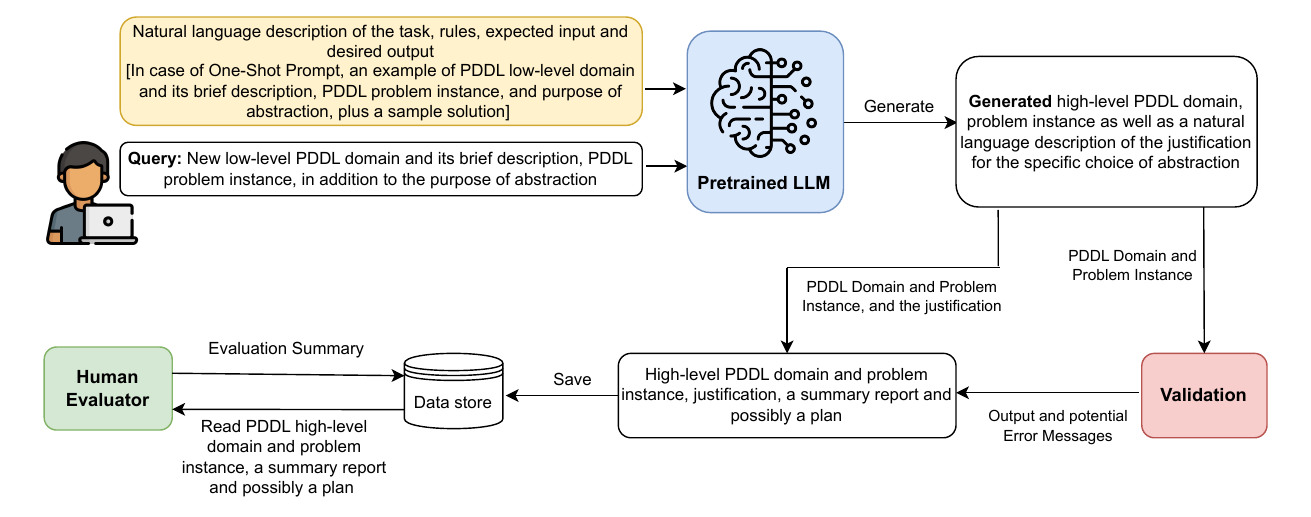}
	\caption{Generating Abstract PDDL Domain and Problem Instance}
	\label{fig:pipeline01}
	\vspace*{1.2em}
\end{figure}

Fig. \ref{fig:pipeline01} shows the outline of our system. Initially, a system prompt (and possibly a user/assistant prompt in case of one-shot in-context learning) is provided to the LLM (GPT-4o) which describes the task, rules to follow, expected input and desired output. 
The user then provides his query which consists of a low-level PDDL domain and its brief description, PDDL
problem instance, and the purpose of abstraction. The LLM in our case is Open AI's GPT-4o and we use Azure OpenAI service to access it.  
After the LLM processes the query, it generates a response which includes a high-level domain and a high-level problem instance in PDDL as well as a natural language description of the justification for the specific choices of abstraction made.  

Our code then extracts the generated high-level PDDL domain and problem instance as .pddl files and sends them to the validation module (which includes VAL and the Fast Downward Planner). 
The validation results and a plan in case of success, or error messages in case of failure, a summary report, in addition to the generated PDDL domain and problem instance  as well as the natural language description of the justification for the specific choices of abstraction
are then saved to a datastore. A human evaluator then reviews the stored information and saves the results of analysis to the datastore.
Note that we don't \emph{formally verify} that the LLM generated high-level domain/problem instance is a sound abstraction and satisfies the constraints on the mapping associated with the purpose of abstraction; this is left for future work. 
 



%
\section{Empirical Evaluation} \label{sec:Eval}

Each of our benchmark examples within each category was run five times.
Tables \ref{table:cat2MAP} to \ref{table:cat4} present the aggregated
results.  The table columns show the following metrics:

\begin{itemize}
	\item Changes Needed (CN): represents the correctness score of
          predicates/actions/types intended to be abstracted
          or eliminated as evaluated by human domain experts (e.g., abstracting types \emph{hotel} and
          \emph{airbnb} to type \emph{accommodation} in Example 1)
	\item Standard Deviation of Changes Needed (CN-SD)
	\item Avoid Unnecessary Changes (AUC): represents the correctness score of predicates/actions/types/objects that were expected to remain unchanged in the high-level domain and problem instance as evaluated by human domain experts (e.g., keeping the type \emph{room} in the abstract domain and problem instance in Example 1)
	\item Standard Deviation of Avoid Unnecessary Changes (AUC-SD)
	\item The last three columns in the table indicate if syntax
          errors were present and whether they were detected by human
          domain experts (HDE), VAL, or the Fast Downward planner (FD).
	
\end{itemize}


\paragraph{Abstraction of Alternative Concrete Actions}
Experimental results for the examples in this category based on zero-shot and one-shot prompting are presented in tables \ref{table:cat2MAP} and \ref{table:cat2EAP} respectively. 

The results show that in general, one-shot prompt achieves better results, except for the case of example \emph{SoftwareDev02}. 
This domain includes three actions: one for creating a wireframe, and
two others for fixing bugs of unmatched parenthesis and wrong
indentation, with a purpose of abstraction which instructs the LLM to
provide an abstraction of the concrete domain such that the high-level
domain abstracts over similar activities in software development. The
expected high-level domain should abstract over the two actions for fixing bugs and retain the action for creating the wireframe. Some common mistakes in the generated results when using one-shot prompting were abstracting the action for creating the wireframe to create a \emph{task}, and abstracting different software bugs and wireframe types as an abstract type \emph{task}.


Our experiments show that changing
the order of action declarations or including additional constructs
irrelevant to the goal of the task do not cause unexpected errors.


In general, the LLM was able to name abstracted
actions/predicates/types reasonably well; an example where naming
needed improvements was in some generated instances of the domain
\emph{DailyPlan01}, where a high-level action abstracting over
alternative choices of concrete actions \emph{orderFood} and
\emph{cook} was named  \emph{eat}. Note that domain constructs may be considered at different levels of detail, e.g., when abstracting types \emph{carrot} and \emph{broccoli}, it is possible to consider high-level types \emph{vegetable}, \emph{plant-based food}, or \emph{food}. In this paper, we do not impose a precise level of abstraction, and leave investigating appropriate levels of abstraction. 


\begin{table}
\captionsetup{font=footnotesize}
\resizebox{0.5\textwidth}{!}{
\begin{tabular}{||c c c c c c c c||} 
 \hline
  & CN & CN-SD & AUC  & AUC-SD & HDE & FD & VAL \\ 
  & (Avg) & & (Avg)  &  &  (Count)  &  (Count) &  (Count) \\ [0.5ex] 
 \hline\hline
  DailyPlan01 & 75.11\% & 13.8 & 100\% & 0 & 2 & 0 & 2 \\ 
 \hline
  DailyPlan02 & 85.07\%  & 5.5 & 98.46\% & 3.08 & 3 & 3 & 3 \\    	
 \hline
  HouseHold01 & 100\% & 0 & 100\% & 0 & 0 & 0 & 0 \\ 
 \hline
 ClothesShop01 & 88.46\% & 14.13 & 74.55\% & 31.18 & 0 & 0 & 0 \\  	
 \hline
 SoftwareDev01 & 94.28\% & 6.29 & 100\% & 0 & 0 & 0 & 0 \\   
 \hline
 SoftwareDev02 & 90.95\% & 4.86 & 97.71\% & 2.8 & 0 & 0 & 0 \\   
 \hline
 SoftwareDev03 & 96\% & 2.42 & 91.11\% & 4.44 & 0 & 0 & 0 \\  
 \hline 
 SoftwareDev04 & 97.88\% & 1.73 & 100\% & 0 & 0 & 0 & 0 \\ 
 \hline
\end{tabular}
}%
\caption{Abstraction of Alternative Concrete Actions - Zero-Shot Prompt}
\label{table:cat2MAP}
\end{table}

\begin{table}
\captionsetup{font=footnotesize}
\resizebox{0.5\textwidth}{!}{
\begin{tabular}{||c c c c c c c c||} 
 \hline
  & CN & CN-SD & AUC  & AUC-SD & HDE & FD & VAL \\ 
  & (Avg) & & (Avg)  &  &  (Count)  &  (Count) &  (Count) \\ [0.5ex] 
 \hline\hline
 DailyPlan01 & 93.19\% & 2.08 & 100\% & 0 & 2 & 2 & 2 \\  
 \hline
 DailyPlan02 & 100\% & 0 & 100\% & 0 & 0 & 0 & 0 \\ 
 \hline
 HouseHold01 & 100\% & 0 & 100\% & 0 & 0 & 0 & 0 \\ 
 \hline
 ClothesShop01 & 100\% & 0 & 100\% & 0 & 0 & 0 & 0 \\ 
 \hline
 SoftwareDev01 & 100\% & 0 & 100\% & 0 & 0 & 0 & 0 \\ 
 \hline
 SoftwareDev02 & 87.14\% & 11.82 & 68\% & 27.01 & 1 & 1 & 0 \\  	
 \hline
 SoftwareDev03 & 98.12\% & 2.31 & 100\% & 0 & 2 & 0 & 2 \\ 
 \hline 
 SoftwareDev04 & 100\% & 0 & 100\% & 0 & 0 & 0 & 0 \\ 
 \hline
\end{tabular}
}%
\caption{Abstraction of Alternative Concrete Actions - One-Shot Prompt}
\label{table:cat2EAP}
\end{table}

\paragraph{Abstraction of Sequences of Concrete Actions}
The results of experiments on the benchmarks of this category (see Table \ref{table:cat3}) show moderate
performance.
A frequent error resulted from the failure to minimize the number of high-level predicates representing conjunctions of low-level predicates.  
On the other hand, in majority of the experiments, low-level actions were abstracted correctly, and planner was able to find the correct high-level plan.

We also noticed that if the concrete actions in a sequence were
closely related, as for instance in the $DeliveryRobot04$ domain which
includes actions \emph{goRestaurant}, \emph{collectOrder},
\emph{goClientHome}, and \emph{giveOrderToClient}, and a purpose of
abstraction that requires abstracting over the detailed steps
of food pickup and detailed steps of delivery may
result in a single high-level action such as \emph{pickUpandDeliver},
when we expect two abstract actions \emph{pickUp} and
\emph{deliver}. This can be fixed by providing a more detailed domain
description which specifically notes two stages in the process, 
and writing a more specific purpose of abstraction which mentions abstracting
over these two stages explicitly.
Note that this problem does not occur if the concrete
actions in the sequence are based on more distinct activities, as
e.g., in \emph{LaptopShop02} which includes actions
\emph{chooseMainBoard}, \emph{chooseMemory}, \emph{enterCCDetails},
and \emph{confirmPayment}; here, the LLM generated two abstract actions, one for customizing the laptop and the other for making payments.

 


Similar to the previous category, we observed that changing the order of action declarations or including additional constructs 
irrelevant to the goal of the task do not cause unexpected errors.

\begin{table}
\captionsetup{font=footnotesize}
\resizebox{0.5\textwidth}{!}{
\begin{tabular}{||c c c c c c c c||} 
 \hline
  & CN & CN-SD & AUC  & AUC-SD & HDE & FD & VAL \\ 
  & (Avg) & & (Avg)  &  &  (Count)  &  (Count) &  (Count) \\ [0.5ex] 
 \hline\hline
  CookFood01 & 79.2\%	 & 6.01 & 88.61\% & 13.41 & 0 & 0 & 0 \\ 
 \hline
  CookFood02 & 72.4\%	& 7.2 & 95.45\% & 9.09 & 0 & 0 & 0 \\ 
 \hline
  CarManufacturing01 & 48.04\%	& 9.71 & 92.68\% & 14.63 & 2 & 2 & 2 \\ 
 \hline
  CleanItem02 & 71.59\%	 & 7.65 & 100\% & 0 & 2 & 1 & 1 \\ 
 \hline
  LaptopShop02 & 65.15\% & 15.15 & 83.64\% & 13.36 & 0 & 0 & 0 \\ 
 \hline
  DeliveryRobot01 & 78\% & 15.86 & 80.47\% & 10.36 & 2 & 1 & 2 \\ 
 \hline
  DeliveryRobot02 & 86.21\%	& 0 & 92.45\% & 0 & 0 & 0 & 0 \\ 
 \hline 
 DeliveryRobot03 & 100\% & 0 & 94.67\%  & 2.67 & 0 & 0 & 0 \\ 
 \hline
 DeliveryRobot04 & 92.55\% & 5.17 & 89.33\% & 9.98 & 1 & 1 & 1 \\ 
 \hline
 RescueRobot02 & 89.92\% & 4.96 & 96.42\% & 5.22 & 0 & 0 & 0 \\ 
 \hline
 RescueRobot03 & 87.5\%	 & 5.24 & 92.54\% & 5.26 & 1 & 1 & 1 \\ 
 \hline
\end{tabular}
}%
\caption{Abstraction of Sequences of Concrete Actions - One-Shot Prompt}
\label{table:cat3}
\end{table}




\paragraph{Abstraction of Action/Predicate Parameters}
The results of experiments on the benchmarks of this category,
where we use zero-shot prompting, are listed in Table
  \ref{table:cat1}. The results indicate that using a zero-shot prompt
  is sufficient for generating mostly correct high-level domain and problem
  instances that are obtained by abstracting away one or more action/predicate parameters.



\begin{table}
\captionsetup{font=footnotesize}
\resizebox{0.5\textwidth}{!}{
\begin{tabular}{||c c c c c c c c||} 
 \hline
  & CN & CN-SD & AUC  & AUC-SD & HDE & FD & VAL \\ 
  & (Avg) & & (Avg)  &  &  (Count)  &  (Count) &  (Count) \\ [0.5ex] 
 \hline\hline
 Email01-0 & 100\% & 0 & 100\% & 0 & 0 & 0 & 0 \\ 
 \hline
 Email01-1 & 100\% & 0 & 100\% & 0 & 0 & 0 & 0 \\ 
 \hline
 Flight01 & 100\% & 0 & 100\% & 0 & 0 & 0 & 0 \\ 
 \hline
 Flight02 & 100\% & 0 & 100\% & 0 & 0 & 0 & 0 \\ 
 \hline
 Library01-1 & 100\% & 0 & 100\% & 0 & 0 & 0 & 0 \\ 
 \hline
 Library01-2 & 96.92\% & 6.15 & 96.10\% & 7.62 & 0 & 0 & 0 \\ 
 \hline
 TechReport01 & 100\% & 0 & 100\% & 0 & 0 & 0 & 0 \\ 
 \hline 
 TechReport02 & 100\% & 0 & 100\% & 0 & 0 & 0 & 0 \\ 
 \hline
 Travel02 & 100\% & 0 & 100\% & 0 & 0 & 0 & 0 \\ 
 \hline
 Travel03 & 100\% & 0 & 100\% & 0 & 0 & 0 & 0 \\ 
 \hline
\end{tabular}
}%
\caption{Abstraction of Action/Predicate Parameters - Zero-Shot Prompt}
\label{table:cat1}
\end{table}


\paragraph{Abstraction of Alternative Sequences of Concrete Actions} Due to the increased complexity of abstractions in this category, the experimental results for the benchmarks (see Table \ref{table:cat4}) indicate a decline in the quality of the generated high-level domains and problem instances, as well as an increase in syntax errors. The majority of mistakes were in generating new predicates which abstract over conjunction, or disjunction of the conjunction of low-level predicates. 
A common syntactic error involved the incorrect use of the keyword ``Or'' with concrete predicates (e.g., as preconditions of actions), instead of creating a high-level predicate which abstracts over them. This occurred despite explicit instructions in the system prompt which requires the generated domains/problem instances to comply with the STRIPS fragment of PDDL, and to avoid using keywords such as 'OR'. 
We also observed that the natural language description of the justification for the specific choices of abstraction which is provided by LLM 
is sometimes inconsistent with the actual generated domain.  

\begin{table}
\captionsetup{font=footnotesize}
\resizebox{0.5\textwidth}{!}{
\begin{tabular}{||c c c c c c c c||} 
 \hline
  & CN & CN-SD & AUC  & AUC-SD & HDE & FD & VAL \\ 
  & (Avg) & & (Avg)  &  &  (Count)  &  (Count) &  (Count) \\ [0.5ex] 
 \hline\hline
 BeveragePreparation01 & 43.92\% & 27.79 & 100\% & 0 & 9 & 1 & 9 \\ 
 \hline
 ClothesShop02 & 65.37\% & 13.59 & 100\% & 0 & 0 & 0 & 0 \\ 
 \hline
 EventPlanning01 & 56.62\% & 1.5 & 96.1\% & 2.46 & 13 & 3 & 14 \\ 
 \hline
 RepairRobot04 & 67.03\% & 5.21 & 87.06\% & 7.8 & 3 & 3 & 3 \\ 
 \hline
 RoomRedesign01 & 69.6\% & 3.32 & 100\% & 0 & 12 & 3 & 12 \\ 
 \hline
 Travel05 & 62.14\% & 11.85 & 100\% & 0 & 10 & 2 & 11 \\ 
 \hline
\end{tabular}
}%
\caption{Abstraction of Alternative Sequences of Concrete Actions}
\label{table:cat4}
\end{table}

\section{Related Work} \label{sec:RelatedWork}



Several authors have studied using LLMs to generate PDDL specifications given natural language descriptions of the task. 
\cite{DBLP:journals/corr/abs-2304-11477} assumes that a PDDL domain description and a contextual example demonstrating the conversion of a natural language problem within the domain into a PDDL problem are provided by human experts, and the LLM is tasked with generation of the PDDL specification of a problem instance (based on the domain) given a natural language description of the problem.
%
\cite{DBLP:journals/corr/abs-2302-05128} studies extracting a planning goal given a natural language instruction, by using zero-shot and few-shot prompting.   
\cite{DBLP:conf/nips/GuanVSK23} follows a strategy where few-shot prompting is used to generate a PDDL representation of a single action at a time, while dynamically updating a list of predicates, and ultimately repeating the process with all the extracted predicates. 
%
\cite{DBLP:conf/icaps/OswaldSK00S24} also uses in-context learning to generate PDDL domains on an action-by-action basis using context examples from other domains. The query in the input prompt includes the allowed predicates that can be used in the definition of that action and a natural language description of the action.
%
%
%
%
The above approaches do not involve generating abstractions of low-level PDDL domains or problem instances. 
%
%
%
To address complex sequential decision-making, \cite{Lie:Choi:Sohn:etal:2024} introduces SkillAct, a prompting method that integrates reusable, task-relevant skill descriptions into prompts. Skills abstract high-level behaviors from agent trajectories (observation-action sequences) and are derived from the LLM's embedded world knowledge via prompting. 
%
%
%
The skills/generated outputs of this approach however are not formalized in PDDL.

%

%
\section{Conclusion and Future Work} \label{sec:ConcFuture}

In this paper, we investigated the feasibility of using an LLM (GPT-4o) to generate high-level PDDL domains and problem instances from low-level PDDL representations, guided by a specified abstraction purpose.

%

In future work, we plan to extend our set of examples and study additional abstraction categories.
We will also consider using the more expressive ADL fragment of PDDL. Building a dataset of planning examples that can be used for fine-tuning an LLM is another avenue for further research. We are also interested in designing other types of prompts that could enhance the generated results. 
We will also examine extending the current abstraction task to incorporate the generation of a refinement mapping so that one can then check that the output high-level model is a sound abstraction of the low-level model relative to this mapping.
%
A significant area for future work lies in the development of (partially) automated validation mechanisms to ensure the correctness of the abstracted domains and problems generated by LLMs. More work on the formalization of the abstraction purpose and automated mechanisms that check the generated models' adherence to it is also indicated.

\section*{Acknowledgements}
This work has been supported by the Natural Sciences and Engineering Research Council of Canada (NSERC) and York University.

\bibliography{citations}

\begin{thebibliography}{37}
\providecommand{\natexlab}[1]{#1}
\providecommand{\url}[1]{\texttt{#1}}
\expandafter\ifx\csname urlstyle\endcsname\relax
  \providecommand{\doi}[1]{doi: #1}\else
  \providecommand{\doi}{doi: \begingroup \urlstyle{rm}\Url}\fi

\bibitem[Bacchus and Kabanza(2000)]{DBLP:journals/ai/BacchusK00}
F.~Bacchus and F.~Kabanza.
\newblock Using temporal logics to express search control knowledge for
  planning.
\newblock \emph{Artificial Intelligence}, 116\penalty0 (1-2):\penalty0
  123--191, 2000.
\newblock \doi{10.1016/S0004-3702(99)00071-5}.

\bibitem[Banihashemi et~al.(2017)Banihashemi, {De Giacomo}, and
  Lesp{\'{e}}rance]{DBLP:conf/aaai/BanihashemiGL17}
B.~Banihashemi, G.~{De Giacomo}, and Y.~Lesp{\'{e}}rance.
\newblock Abstraction in situation calculus action theories.
\newblock In \emph{Proceedings of the Thirty-First {AAAI} Conference on
  Artificial Intelligence}, pages 1048--1055. {AAAI} Press, 2017.

\bibitem[Banihashemi et~al.(2025)Banihashemi, Giacomo, and
  Lesp{\'{e}}rance]{DBLP:journals/ai/BanihashemiGL25}
B.~Banihashemi, G.~D. Giacomo, and Y.~Lesp{\'{e}}rance.
\newblock Abstracting situation calculus action theories.
\newblock \emph{Artif. Intell.}, 348:\penalty0 104407, 2025.
\newblock \doi{10.1016/J.ARTINT.2025.104407}.
\newblock URL \url{https://doi.org/10.1016/j.artint.2025.104407}.

\bibitem[Brown et~al.(2020)Brown, Mann, Ryder, Subbiah, Kaplan, Dhariwal,
  Neelakantan, Shyam, Sastry, Askell, Agarwal, Herbert{-}Voss, Krueger,
  Henighan, Child, Ramesh, Ziegler, Wu, Winter, Hesse, Chen, Sigler, Litwin,
  Gray, Chess, Clark, Berner, McCandlish, Radford, Sutskever, and
  Amodei]{DBLP:journals/corr/abs-2005-14165}
T.~B. Brown, B.~Mann, N.~Ryder, M.~Subbiah, J.~Kaplan, P.~Dhariwal,
  A.~Neelakantan, P.~Shyam, G.~Sastry, A.~Askell, S.~Agarwal,
  A.~Herbert{-}Voss, G.~Krueger, T.~Henighan, R.~Child, A.~Ramesh, D.~M.
  Ziegler, J.~Wu, C.~Winter, C.~Hesse, M.~Chen, E.~Sigler, M.~Litwin, S.~Gray,
  B.~Chess, J.~Clark, C.~Berner, S.~McCandlish, A.~Radford, I.~Sutskever, and
  D.~Amodei.
\newblock Language models are few-shot learners.
\newblock \emph{CoRR}, abs/2005.14165, 2020.

\bibitem[Chen and Bercher(2021)]{DBLP:conf/aips/ChenB21}
D.~Chen and P.~Bercher.
\newblock Fully observable nondeterministic {HTN} planning - formalisation and
  complexity results.
\newblock In \emph{Proceedings of the Thirty-First International Conference on
  Automated Planning and Scheduling, {ICAPS} 2021}, pages 74--84. {AAAI} Press,
  2021.

\bibitem[Cla{\ss}en et~al.(2007)Cla{\ss}en, Eyerich, Lakemeyer, and
  Nebel]{DBLP:conf/ijcai/ClassenELN07}
J.~Cla{\ss}en, P.~Eyerich, G.~Lakemeyer, and B.~Nebel.
\newblock Towards an integration of golog and planning.
\newblock In \emph{{IJCAI}}, pages 1846--1851, 2007.

\bibitem[{De Giacomo} et~al.(2000){De Giacomo}, Lesp{\'{e}}rance, and
  Levesque]{DBLP:journals/ai/GiacomoLL00}
G.~{De Giacomo}, Y.~Lesp{\'{e}}rance, and H.~J. Levesque.
\newblock {ConGolog}, a concurrent programming language based on the situation
  calculus.
\newblock \emph{Artificial Intelligence}, 121\penalty0 (1-2):\penalty0
  109--169, 2000.
\newblock \doi{10.1016/S0004-3702(00)00031-X}.

\bibitem[{De Giacomo} et~al.(2012){De Giacomo}, Lesp{\'{e}}rance, and
  Muise]{DBLP:conf/aamas/GiacomoLM12}
G.~{De Giacomo}, Y.~Lesp{\'{e}}rance, and C.~J. Muise.
\newblock On supervising agents in situation-determined {ConGolog}.
\newblock In \emph{International Conference on Autonomous Agents and Multiagent
  Systems, {AAMAS} 2012}, pages 1031--1038. {IFAAMAS}, 2012.

\bibitem[Dong et~al.(2025)Dong, Shi, Zeng, and Liu]{DBLP:conf/aaai/DongSZL25}
H.~Dong, Z.~Shi, H.~Zeng, and Y.~Liu.
\newblock An automatic sound and complete abstraction method for generalized
  planning with baggable types.
\newblock In \emph{AAAI-25, Sponsored by the Association for the Advancement of
  Artificial Intelligence}, pages 14875--14884. {AAAI} Press, 2025.
\newblock \doi{10.1609/AAAI.V39I14.33631}.

\bibitem[Dong et~al.(2024)Dong, Li, Dai, Zheng, Wu, Chang, Sun, Xu, Li, and
  Sui]{DBLP:journals/corr/abs-2301-00234}
Q.~Dong, L.~Li, D.~Dai, C.~Zheng, Z.~Wu, B.~Chang, X.~Sun, J.~Xu, L.~Li, and
  Z.~Sui.
\newblock A survey on in-context learning.
\newblock \emph{CoRR}, abs/2301.00234, 2024.
\newblock \doi{10.48550/ARXIV.2301.00234}.

\bibitem[Erol et~al.(1996)Erol, Hendler, and Nau]{DBLP:journals/amai/ErolHN96}
K.~Erol, J.~A. Hendler, and D.~S. Nau.
\newblock Complexity results for {HTN} planning.
\newblock \emph{Annals of Mathematics and Artificial Intelligence}, 18\penalty0
  (1):\penalty0 69--93, 1996.
\newblock \doi{10.1007/BF02136175}.

\bibitem[Fang et~al.(2025)Fang, Wang, Chen, Luo, Cui, and
  Guan]{DBLP:conf/aaai/FangWCLCG25}
L.~Fang, X.~Wang, Z.~Chen, K.~Luo, Z.~Cui, and Q.~Guan.
\newblock A syntactic approach to computing complete and sound abstraction in
  the situation calculus.
\newblock In \emph{AAAI-25, Sponsored by the Association for the Advancement of
  Artificial Intelligence}, pages 14911--14921. {AAAI} Press, 2025.
\newblock \doi{10.1609/AAAI.V39I14.33635}.

\bibitem[Ghallab et~al.(2004)Ghallab, Nau, and
  Traverso]{DBLP:books/daglib/0014222}
M.~Ghallab, D.~S. Nau, and P.~Traverso.
\newblock \emph{Automated planning - theory and practice}.
\newblock Elsevier, 2004.
\newblock ISBN 978-1-55860-856-6.

\bibitem[Guan et~al.(2023)Guan, Valmeekam, Sreedharan, and
  Kambhampati]{DBLP:conf/nips/GuanVSK23}
L.~Guan, K.~Valmeekam, S.~Sreedharan, and S.~Kambhampati.
\newblock Leveraging pre-trained large language models to construct and utilize
  world models for model-based task planning.
\newblock In \emph{Advances in Neural Information Processing Systems 36: Annual
  Conference on Neural Information Processing Systems 2023, NeurIPS 2023},
  2023.

\bibitem[He et~al.(2023)He, Huang, Xiao, and Liu]{DBLP:conf/acl/0001HX023}
W.~He, C.~Huang, Z.~Xiao, and Y.~Liu.
\newblock Exploring the capacity of pretrained language models for reasoning
  about actions and change.
\newblock In \emph{Proceedings of the 61st Annual Meeting of the Association
  for Computational Linguistics (Volume 1: Long Papers), {ACL} 2023}, pages
  4629--4643. Association for Computational Linguistics, 2023.
\newblock \doi{10.18653/V1/2023.ACL-LONG.255}.

\bibitem[Helmert(2006)]{DBLP:journals/jair/Helmert06}
M.~Helmert.
\newblock The fast downward planning system.
\newblock \emph{Journal of Artificial Intelligence Research}, 26:\penalty0
  191--246, 2006.
\newblock \doi{10.1613/JAIR.1705}.

\bibitem[Howey et~al.(2004)Howey, Long, and Fox]{DBLP:conf/ictai/HoweyLF04}
R.~Howey, D.~Long, and M.~Fox.
\newblock {VAL:} automatic plan validation, continuous effects and mixed
  initiative planning using {PDDL}.
\newblock In \emph{16th {IEEE} International Conference on Tools with
  Artificial Intelligence {(ICTAI} 2004)}, pages 294--301. {IEEE} Computer
  Society, 2004.
\newblock \doi{10.1109/ICTAI.2004.120}.

\bibitem[{Hurst} et~al.(2024)]{DBLP:journals/corr/abs-2410-21276}
A.~{Hurst} et~al.
\newblock Gpt-4o system card.
\newblock \emph{CoRR}, abs/2410.21276, 2024.
\newblock \doi{10.48550/ARXIV.2410.21276}.

\bibitem[Kong et~al.(2024)Kong, Zhao, Chen, Li, Qin, Sun, Zhou, Wang, and
  Dong]{DBLP:conf/naacl/KongZCLQSZWD24}
A.~Kong, S.~Zhao, H.~Chen, Q.~Li, Y.~Qin, R.~Sun, X.~Zhou, E.~Wang, and
  X.~Dong.
\newblock Better zero-shot reasoning with role-play prompting.
\newblock In \emph{Proceedings of the 2024 Conference of the North American
  Chapter of the Association for Computational Linguistics: Human Language
  Technologies (Volume 1: Long Papers), {NAACL} 2024}, pages 4099--4113.
  Association for Computational Linguistics, 2024.
\newblock \doi{10.18653/V1/2024.NAACL-LONG.228}.

\bibitem[Li(2023)]{DBLP:conf/ranlp/Li23}
Y.~Li.
\newblock A practical survey on zero-shot prompt design for in-context
  learning.
\newblock In \emph{Proceedings of the 14th International Conference on Recent
  Advances in Natural Language Processing, {RANLP} 2023}, pages 641--647.
  {INCOMA} Ltd., Shoumen, Bulgaria, 2023.

\bibitem[Liu et~al.(2024)Liu, Choi, Sohn, Fu, Kim, Kim, Wang, Yoo, and
  Lee]{Lie:Choi:Sohn:etal:2024}
A.~Z. Liu, J.~Choi, S.~Sohn, Y.~Fu, J.~Kim, D.-K. Kim, X.~Wang, J.~Yoo, and
  H.~Lee.
\newblock {SkillAct}: Using skill abstractions improves {LLM} agents.
\newblock In \emph{ICML 2024 Workshop on {LLMs} and Cognition}, 2024.

\bibitem[Liu et~al.(2023)Liu, Jiang, Zhang, Liu, Zhang, Biswas, and
  Stone]{DBLP:journals/corr/abs-2304-11477}
B.~Liu, Y.~Jiang, X.~Zhang, Q.~Liu, S.~Zhang, J.~Biswas, and P.~Stone.
\newblock {LLM+P:} empowering large language models with optimal planning
  proficiency.
\newblock \emph{CoRR}, abs/2304.11477, 2023.
\newblock \doi{10.48550/ARXIV.2304.11477}.

\bibitem[Luo(2023)]{DBLP:conf/aaai/Luo23}
K.~Luo.
\newblock Automated verification of propositional agent abstraction for
  classical planning via {CTLK} model checking.
\newblock In \emph{Thirty-Seventh {AAAI} Conference on Artificial Intelligence,
  {AAAI} 2023}, pages 6475--6482. {AAAI} Press, 2023.

\bibitem[Luo et~al.(2020)Luo, Liu, Lesp{\'{e}}rance, and Lin]{LuoECAI20}
K.~Luo, Y.~Liu, Y.~Lesp{\'{e}}rance, and Z.~Lin.
\newblock Agent abstraction via forgetting in the situation calculus.
\newblock In \emph{{ECAI} 2020 - 24th European Conference on Artificial
  Intelligence}, volume 325 of \emph{Frontiers in Artificial Intelligence and
  Applications}, pages 809--816. {IOS} Press, 2020.

\bibitem[McCarthy and Hayes(1969)]{McCarthy1969:AI}
J.~McCarthy and P.~J. Hayes.
\newblock Some philosophical problems from the standpoint of artificial
  intelligence.
\newblock \emph{Machine Intelligence}, 4:\penalty0 463--502, 1969.

\bibitem[McDermott et~al.(1998)McDermott, Ghallab, Howe, Knoblock, Ram, Veloso,
  Weld, and Wilkins]{McDermot:etal:1998}
D.~McDermott, M.~Ghallab, A.~Howe, C.~Knoblock, A.~Ram, M.~Veloso, D.~Weld, and
  D.~Wilkins.
\newblock {PDDL} – the planning domain definition language.
\newblock Technical Report CVC TR-98-003/DCS TR-1165, Yale Center for
  Computational Vision and Control, 1998.

\bibitem[Milner(1971)]{DBLP:conf/ijcai/Milner71}
R.~Milner.
\newblock An algebraic definition of simulation between programs.
\newblock In \emph{Proceedings of the 2nd International Joint Conference on
  Artificial Intelligence}, pages 481--489. William Kaufmann, 1971.

\bibitem[Milner(1989)]{DBLP:books/daglib/0067019}
R.~Milner.
\newblock \emph{Communication and concurrency}.
\newblock {PHI} Series in computer science. Prentice Hall, 1989.
\newblock ISBN 978-0-13-115007-2.

\bibitem[Oswald et~al.(2024)Oswald, Srinivas, Kokel, Lee, Katz, and
  Sohrabi]{DBLP:conf/icaps/OswaldSK00S24}
J.~T. Oswald, K.~Srinivas, H.~Kokel, J.~Lee, M.~Katz, and S.~Sohrabi.
\newblock Large language models as planning domain generators.
\newblock In \emph{Proceedings of the Thirty-Fourth International Conference on
  Automated Planning and Scheduling, {ICAPS} 2024}, pages 423--431. {AAAI}
  Press, 2024.
\newblock \doi{10.1609/ICAPS.V34I1.31502}.

\bibitem[Pednault(1989)]{DBLP:conf/kr/Pednault89}
E.~P.~D. Pednault.
\newblock {ADL:} exploring the middle ground between {STRIPS} and the situation
  calculus.
\newblock In \emph{Proceedings of the 1st International Conference on
  Principles of Knowledge Representation and Reasoning (KR'89)}, pages
  324--332. Morgan Kaufmann, 1989.

\bibitem[Reiter(2001)]{Reiter01-Book}
R.~Reiter.
\newblock \emph{Knowledge in Action. {L}ogical Foundations for Specifying and
  Implementing Dynamical Systems}.
\newblock The MIT Press, 2001.

\bibitem[Seegebarth et~al.(2012)Seegebarth, M{\"{u}}ller, Schattenberg, and
  Biundo]{DBLP:conf/aips/SeegebarthMSB12}
B.~Seegebarth, F.~M{\"{u}}ller, B.~Schattenberg, and S.~Biundo.
\newblock Making hybrid plans more clear to human users - {A} formal approach
  for generating sound explanations.
\newblock In \emph{Proceedings of the Twenty-Second International Conference on
  Automated Planning and Scheduling, {ICAPS} 2012}. {AAAI}, 2012.

\bibitem[Sutton et~al.(1999)Sutton, Precup, and
  Singh]{DBLP:journals/ai/SuttonPS99}
R.~S. Sutton, D.~Precup, and S.~Singh.
\newblock Between {MDPs and Semi-MDPs}: {A} framework for temporal abstraction
  in reinforcement learning.
\newblock \emph{Artificial Intelligence}, 112\penalty0 (1-2):\penalty0
  181--211, 1999.
\newblock \doi{10.1016/S0004-3702(99)00052-1}.

\bibitem[Valmeekam et~al.(2024)Valmeekam, Stechly, Gundawar, and
  Kambhampati]{DBLP:journals/corr/abs-2410-02162}
K.~Valmeekam, K.~Stechly, A.~Gundawar, and S.~Kambhampati.
\newblock Planning in strawberry fields: Evaluating and improving the planning
  and scheduling capabilities of {LRM} o1.
\newblock \emph{CoRR}, abs/2410.02162, 2024.
\newblock \doi{10.48550/ARXIV.2410.02162}.

\bibitem[Wei et~al.(2022)Wei, Wang, Schuurmans, Bosma, Ichter, Xia, Chi, Le,
  and Zhou]{DBLP:conf/nips/Wei0SBIXCLZ22}
J.~Wei, X.~Wang, D.~Schuurmans, M.~Bosma, B.~Ichter, F.~Xia, E.~H. Chi, Q.~V.
  Le, and D.~Zhou.
\newblock Chain-of-thought prompting elicits reasoning in large language
  models.
\newblock In \emph{Advances in Neural Information Processing Systems 35: Annual
  Conference on Neural Information Processing Systems 2022, NeurIPS 2022},
  2022.

\bibitem[Xie et~al.(2023)Xie, Yu, Zhu, Bai, Gong, and
  Soh]{DBLP:journals/corr/abs-2302-05128}
Y.~Xie, C.~Yu, T.~Zhu, J.~Bai, Z.~Gong, and H.~Soh.
\newblock Translating natural language to planning goals with large-language
  models.
\newblock \emph{CoRR}, abs/2302.05128, 2023.
\newblock \doi{10.48550/ARXIV.2302.05128}.

\bibitem[Zhao et~al.(2023)Zhao, Zhou, Li, Tang, Wang, Hou, Min, Zhang, Zhang,
  Dong, Du, Yang, Chen, Chen, Jiang, Ren, Li, Tang, Liu, Liu, Nie, and
  Wen]{DBLP:journals/corr/abs-2303-18223}
W.~X. Zhao, K.~Zhou, J.~Li, T.~Tang, X.~Wang, Y.~Hou, Y.~Min, B.~Zhang,
  J.~Zhang, Z.~Dong, Y.~Du, C.~Yang, Y.~Chen, Z.~Chen, J.~Jiang, R.~Ren, Y.~Li,
  X.~Tang, Z.~Liu, P.~Liu, J.~Nie, and J.~Wen.
\newblock A survey of large language models.
\newblock \emph{CoRR}, abs/2303.18223, 2023.
\newblock \doi{10.48550/ARXIV.2303.18223}.

\end{thebibliography}

\clearpage
\appendix
\section{Appendix - Using Large Language Models for Abstraction of Planning Domains}

\subsection{Categories} \label{subSec:AppExamples}

\subsubsection{Abstraction of Alternative Concrete Actions} \label{subsubSec:AbsAltLLProblem}

The low-level domain:

\begin{lstlisting} [numbers=none, 
                    basicstyle=\scriptsize\sffamily, 
										xleftmargin=0.2em, 
										columns=fixed,       
                    keepspaces=true,
										breaklines=true,
										postbreak=\mbox{\textcolor{blue}{$\hookrightarrow$}\space\space},										
										label={lst:Cat2TravelLLApp}]
(define (domain travelArrange01_LL)
(:requirements :strips :typing)
(:types 
    hotel airbnb room flight trainRide seat - object)
(:predicates
    (booked_hotel  ?r -room  ?h - hotel)
    (booked_airbnb  ?r - room ?ab - airbnb)
    (available_room_hotel  ?r  - room ?h - hotel)
    (available_room_airbnb  ?r - room ?ab - airbnb)
    (bookedHotelOrAirbnb)
    (available_seat_flight ?s - seat ?f - flight)
    (available_seat_trainRide ?s -seat ?t - trainRide)
    (booked_flight ?s - seat ?f - flight)
    (booked_trainRide ?s - seat ?t - trainRide)
    (bookedFlightOrTrainRide))
(:action book_hotel
    :parameters (?h - hotel ?r - room)
    :precondition (available_room_hotel ?r ?h)
    :effect (and (booked_hotel ?r ?h) 
                 (not (available_room_hotel ?r ?h))
                 (bookedHotelOrAirbnb)))    
(:action book_airbnb
    :parameters (?ab - airbnb ?r - room)
    :precondition (available_room_airbnb ?r ?ab)                          
    :effect (and (booked_airbnb ?r ?ab) 
                 (not (available_room_airbnb ?r ?ab))
                 (bookedHotelOrAirbnb)))      
(:action book_flight
    :parameters (?f - flight ?s - seat)
    :precondition (available_seat_flight ?s ?f)                          
    :effect (and (booked_flight ?s ?f) 
                 (not (available_seat_flight ?s ?f))
                 (bookedFlightOrTrainRide)))     
(:action book_trainRide
    :parameters (?t - trainRide ?s - seat)
    :precondition (available_seat_trainRide ?s ?t)                          
    :effect (and (booked_trainRide ?s ?t) 
                 (not (available_seat_trainRide ?s ?t))
                 (bookedFlightOrTrainRide))))
\end{lstlisting}

The low-level problem instance:

\noindent

\begin{lstlisting} [numbers=none, 
                    basicstyle=\scriptsize\sffamily, 
										xleftmargin=0.2em, 
										columns=fixed,       
                    keepspaces=true,
										breaklines=true,
										postbreak=\mbox{\textcolor{blue}{$\hookrightarrow$}\space\space},
										label={lst:Cat2TrPrLL}]

(define (problem travelArrange01Problem1_LL)
(:domain travelArrange01_LL)

(:objects 
    room1 room2 room3 room4 - room
    hotel1 - hotel
    airbnb1 - airbnb
    seat1 seat2 seat3 seat4 - seat
    flight1 - flight
    trainRide1 - trainRide)     
(:init 
    (available_room_hotel  room1 hotel1)
    (available_room_hotel  room2 hotel1)
    (available_room_airbnb  room3 airbnb1)  
    (available_room_airbnb  room4 airbnb1)
    (available_seat_flight  seat1 flight1)
    (available_seat_flight  seat2 flight1)
    (available_seat_trainRide  seat3 trainRide1) 
    (available_seat_trainRide  seat4 trainRide1))
(:goal 
   (and (bookedFlightOrTrainRide) 
        (bookedHotelOrAirbnb)))   
)
\end{lstlisting}%

The following high-level problem domain and problem instance are based on the purpose of abstraction ``provide an abstraction of the concrete domain such that the high-level domain abstracts over booking various kinds of transportation and accommodation''.

The high-level domain:

\begin{lstlisting} [numbers=none, 
                    basicstyle=\scriptsize\sffamily, 
										xleftmargin=0.2em, 
										columns=fixed,       
                    keepspaces=true,
										breaklines=true,
										postbreak=\mbox{\textcolor{blue}{$\hookrightarrow$}\space\space},										
										label={lst:Cat2TravelHLApp}]
(define (domain travelArrange01_HL)
(:requirements :strips :typing)
(:types accommodation room transportation seat 
 - object)
(:predicates
    (booked_accommodation  ?r - room ?a - accommodation)
    (available_room  ?r - room ?a - accommodation)
    (doneBookingAccommodation)
    (available_seat ?s - seat ?tp - transportation)
    (booked_transportation ?s - seat ?tp - transportation)
    (doneBookingTransportation))
(:action book_accommodation
    :parameters (?a - accommodation ?r - room)
    :precondition (available_room ?r ?a)                          
    :effect (and (booked_accommodation ?r ?a) 
                 (not (available_room ?r ?a))
                 (doneBookingAccommodation)))      
(:action book_transportation
   :parameters (?tp - transportation ?s - seat)
   :precondition (available_seat ?s ?tp)                          
   :effect (and (booked_transportation ?s ?tp) 
                (not (available_seat ?s ?tp))
                (doneBookingTransportation))))
\end{lstlisting}

The high-level problem instance:

\begin{lstlisting} [numbers=none, 
                    basicstyle=\scriptsize\sffamily, 
										xleftmargin=0.2em, 
										columns=fixed,       
                    keepspaces=true,
										breaklines=true,
										postbreak=\mbox{\textcolor{blue}{$\hookrightarrow$}\space\space},
										label={lst:Cat2TrPrHL}]

(define (problem travelArrange01Problem1_HL)
(:domain travelArrange01_HL)

(:objects 
    room1 room2 room3 room4 - room
    hotel1 airbnb1 - accommodation
    seat1 seat2 seat3 seat4 - seat
    flight1 trainRide1 - transportation )     
(:init 
    (available_room  room1 hotel1)
    (available_room  room2 hotel1)
    (available_room  room3 airbnb1)  
    (available_room  room4 airbnb1)
    (available_seat  seat1 flight1)
    (available_seat  seat2 flight1)
    (available_seat  seat3 trainRide1) 
    (available_seat  seat4 trainRide1) )    
(:goal 
   (and (doneBookingAccommodation) 
        (doneBookingTransportation) ) )   
)
\end{lstlisting}%


\subsubsection{Abstraction of Sequences of Concrete Actions} \label{subsubSec:AbsSeqLLProblem}

The low-level domain:

\noindent
\begin{lstlisting} [numbers=none, 
                    basicstyle=\scriptsize\sffamily, 
										xleftmargin=0.2em, 
										columns=fixed,       
                    keepspaces=true,
										breaklines=true,
										postbreak=\mbox{\textcolor{blue}{$\hookrightarrow$}\space\space},
										label={lst:Cat3CloudLLAp}]

(define (domain cloudApps01_LL)
(:requirements :strips :typing)
(:types file userName passWord - object)
(:predicates
    (authenticated_userName ?u - userName)
    (authenticated_passWord ?p - passWord)
    (openedFileInEditor ?f - file)
    (hasEditPermission ?u - userName ?f - file)
    (closed_file ?f - file)
    (valid_userName  ?u - userName)
    (valid_passWord ?p - passWord)
    (authenticUserPassword ?u - userName ?p - passWord)     
    (changedFileContent ?f - file))
 (:action enter_UserName
        :parameters (?u - userName)
        :precondition (valid_userName ?u)
        :effect (authenticated_userName ?u))
 (:action enter_passWord
        :parameters (?u - userName ?p - passWord)
        :precondition (and (valid_passWord ?p)
                           (authenticUserPassword ?u ?p)                     
                           (authenticated_userName ?u))                          
        :effect (authenticated_passWord ?p))      
 (:action openFileInEditor
        :parameters (?f - file ?p - passWord ?u - userName)
        :precondition (and (closed_file ?f)
                           (hasEditPermission ?u ?f) 
                           (authenticated_passWord ?p))                          
        :effect (and (openedFileInEditor ?f)
                     (not(closed_file ?f)) ))   
 (:action changeFileContent
        :parameters (?f - file)
        :precondition (openedFileInEditor ?f)                          
        :effect (changedFileContent ?f)))     
\end{lstlisting}%

The low-level problem instance:

\noindent
\begin{lstlisting} [numbers=none, 
                    basicstyle=\scriptsize\sffamily, 
										xleftmargin=0.2em, 
										columns=fixed,       
                    keepspaces=true,
										breaklines=true,
										postbreak=\mbox{\textcolor{blue}{$\hookrightarrow$}\space\space},
										label={lst:Cat3ClPrLL}]

(define (problem cloudApps01Problem1_LL)
(:domain cloudApps01_LL)

(:objects 
    file1  - file
    user1 user2 - userName
    pw1 pw2 - passWord)     
(:init 
    (closed_file file1)
    (valid_userName  user1)
    (valid_passWord pw1) 
    (authenticUserPassword user1 pw1)
    (authenticUserPassword user2 pw2)
    (hasEditPermission user1 file1))    
(:goal 
   (changedFileContent file1))   
)
\end{lstlisting}%

The following high-level problem domain and instance are based on the purpose of abstraction ``provide an abstraction of the concrete domain such that the high-level domain abstracts over detailed steps of logging in and editing a file on cloud''.

The high-level domain:

\begin{lstlisting} [numbers=none, 
                    basicstyle=\scriptsize\sffamily, 
										xleftmargin=0.2em, 
										columns=fixed,       
                    keepspaces=true,
										breaklines=true,
										postbreak=\mbox{\textcolor{blue}{$\hookrightarrow$}\space\space},
										label={lst:Cat3CloudHLA}]

(define (domain cloudApps01_HL)
(:requirements :strips :typing)
(:types file userName passWord - object)
(:predicates
     (logged_in ?u - userName ?p - passWord)
     (edited_file  ?f - file)
     (closed_file ?f - file)
     (hasEditPermission ?u - userName ?f - file)
     (valid_credentials  ?u - userName ?p - passWord))
(:action login
     :parameters (?u - userName ?p - passWord)
     :precondition (valid_credentials ?u ?p)
     :effect (logged_in ?u ?p))
(:action edit_file
     :parameters (?f - file ?p - passWord ?u - userName)
     :precondition (and (closed_file ?f)
                        (hasEditPermission ?u ?f)
                        (logged_in ?u ?p) )                          
      :effect (and (edited_file ?f) 
                   (not(closed_file ?f)) ) ) ) 
\end{lstlisting}

The high-level problem instance:

\noindent
\begin{lstlisting} [numbers=none, 
                    basicstyle=\scriptsize\sffamily, 
										xleftmargin=0.2em, 
										columns=fixed,       
                    keepspaces=true,
										breaklines=true,
										postbreak=\mbox{\textcolor{blue}{$\hookrightarrow$}\space\space},
										label={lst:Cat3ClPrHL}]

(define (problem cloudApps01Problem1_HL)
(:domain cloudApps01_HL)

(:objects 
    file1 - file
    user1 user2 - userName
    pw1 pw2 - passWord) 		
(:init 
    (closed_file  file1)
    (hasEditPermission user1 file1)
    (valid_credentials  user1 pw1) 
    (valid_credentials  user2 pw2))		
(:goal 
   (edited_file file1) )   
)
\end{lstlisting}%


\subsubsection{Abstraction of Action/Predicate Parameters} \label{subsubSec:AbsFeatureLLProblem}

The low-level domain:

\noindent
\begin{lstlisting} [numbers=none, 
                    basicstyle=\scriptsize\sffamily, 
										xleftmargin=0.2em, 
										columns=fixed,       
                    keepspaces=true,
										breaklines=true,
										postbreak=\mbox{\textcolor{blue}{$\hookrightarrow$}\space\space},
										label={lst:Cat1HBLLA}]

(define (domain travelArrange02_LL) 
(:requirements  :strips  :typing) 
(:types hotel r_view room r_type - object)
(:predicates 
  (booked_hotel  ?rt - r_type ?h - hotel  ?rv - r_view) 
  (available_room_hotel   ?r  -  room  ?h - hotel) 
  (roomType  ?r - room ?rt - r_type) 
  (roomView  ?r - room ?rv - r_view) 
  (difRoomType ?rt1 - r_type ?rt2 - r_type) )
(:action book_hotel 
  :parameters (?h - hotel ?r - room ?rv - r_view ?rt - r_type) 
  :precondition (and (available_room_hotel ?r ?h) 
                     (roomType ?r ?rt)  
                     (roomView ?r ?rv))  
  :effect (and (booked_hotel ?rt ?h ?rv) 
               (not (available_room_hotel ?r ?h))))      
(:action  change_RoomType  
   :parameters (?r - room ?rt1 ?rt2 - r_type) 
   :precondition (and (roomType ?r ?rt1) 
                      (difRoomType ?rt1 ?rt2)                      
   :effect (and (roomType ?r ?rt2) 
                (not (roomType ?r ?rt1))))   
\end{lstlisting}%

The low-level problem instance:

\noindent
\begin{lstlisting} [numbers=none, 
                    basicstyle=\scriptsize\sffamily, 
										xleftmargin=0.2em, 
										columns=fixed,       
                    keepspaces=true,
										breaklines=true,
										postbreak=\mbox{\textcolor{blue}{$\hookrightarrow$}\space\space},										
										label={lst:Cat1HBPrLL}]
(define (problem travelArrange02Problem1_LL)
(:domain travelArrange02_LL)

(:objects 
    room1 room2 room3  - room
    hotel1 hotel2 - hotel
    single double - r_type
    oceanView gardenView - r_view)     
(:init 
    (available_room_hotel  room1 hotel1)
    (available_room_hotel  room2 hotel1)
    (available_room_hotel  room3 hotel2)
    (roomType room1  single)
    (roomType room2  double)
    (roomType room3  single)
    (roomView room1  oceanView)
    (roomView room2  gardenView)
    (roomView room3  oceanView)
    (difRoomType single double)
    (difRoomType double single))    
(:goal 
   (booked_hotel single hotel1 oceanView)))
\end{lstlisting}%

The following high-level domain and problem instance are based on the purpose of abstraction ``provide an abstraction of the concrete domain such that the high-level domain must not include information about room view''. 

The high-level domain:

\noindent
\begin{lstlisting} [numbers=none, 
                    basicstyle=\scriptsize\sffamily, 
										xleftmargin=0.2em, 
										columns=fixed,       
                    keepspaces=true,
										breaklines=true,
										postbreak=\mbox{\textcolor{blue}{$\hookrightarrow$}\space\space},										
										label={lst:Cat1HBHLRVApx}]
(define (domain travelArrange02_HL)
(:requirements strips :typing)
(:types hotel room r_type - object)
(:predicates
    (booked_hotel   ?rt - r_type ?h - hotel)
    (available_room_hotel  ?r  - room ?h - hotel)
    (roomType ?r - room ?rt - r_type)
    (difRoomType ?rt1 - r_type ?rt2 - r_type))
(:action book_hotel
    :parameters (?h - hotel ?r - room ?rt - r_type)
    :precondition (and (available_room_hotel ?r ?h)
                       (roomType ?r ?rt))
    :effect (and (booked_hotel ?rt ?h) 
                 (not (available_room_hotel ?r ?h))))    
(:action change_RoomType
    :parameters (?r - room ?rt1 ?rt2 - r_type)
    :precondition (and (roomType ?r ?rt1)
                       (difRoomType ?rt1 ?rt2))                          
    :effect (and (roomType ?r ?rt2) 
                 (not (roomType ?r ?rt1))))
\end{lstlisting}

The high-level problem instance:

\noindent
\begin{lstlisting} [numbers=none, 
                    basicstyle=\scriptsize\sffamily, 
										xleftmargin=0.2em, 
										columns=fixed,       
                    keepspaces=true,
										breaklines=true,
										postbreak=\mbox{\textcolor{blue}{$\hookrightarrow$}\space\space},
										label={lst:Cat1HBPrHL}]
(define (problem travelArrange02Problem1_HL)
(:domain travelArrange02_HL)

(:objects 
    room1 room2 room3  - room_HL
    single double - r_type_HL
    hotel1 hotel2 - hotel_HL)     
(:init 
    (available_room_hotel_HL  room1 hotel1)
    (available_room_hotel_HL  room2 hotel1)
    (available_room_hotel_HL  room3 hotel2)
    (roomType_HL room1  single)
    (roomType_HL room2  double)
    (roomType_HL room3  single)
    (difRoomType_HL single double)
    (difRoomType_HL double single))    
(:goal 
   (booked_hotel_HL single hotel1)))
\end{lstlisting}

The following high-level domain and problem instance are based on the purpose of abstraction ``provide an abstraction of the concrete domain such that the high-level domain must not include information about room type''. 

High-level domain:

\noindent
\begin{lstlisting} [numbers=none, 
                    basicstyle=\scriptsize\sffamily, 
										xleftmargin=0.2em, 
										columns=fixed,       
                    keepspaces=true,
										breaklines=true,
										postbreak=\mbox{\textcolor{blue}{$\hookrightarrow$}\space\space},
										label={lst:Cat1HBHLRTAp}]
(define (domain travelArrange03_HL)
(:requirements :strips :typing)
(:types hotel r_view room - object)
(:predicates
    (booked_hotel ?h - hotel ?rv - r_view)
    (available_room_hotel  ?r  - room ?h - hotel)
    (roomView ?r - room ?rv - r_view))
(:action book_hotel
    :parameters (?h - hotel ?r - room ?rv - r_view)
    :precondition (and (available_room_hotel ?r ?h)
                       (roomView ?r ?rv))    
    :effect (and (booked_hotel ?h ?rv) 
                 (not (available_room_hotel ?r ?h)))))
\end{lstlisting}

High-level problem instance:

\begin{lstlisting} [numbers=none, 
                    basicstyle=\scriptsize\sffamily, 
										xleftmargin=0.2em, 
										columns=fixed,       
                    keepspaces=true,
										breaklines=true,
										postbreak=\mbox{\textcolor{blue}{$\hookrightarrow$}\space\space},
										label={lst:Cat1HBHLRTPrAp}]										
(define (problem travelArrange03Problem1_LL)
(:domain travelArrange03_LL)

(:objects 
    room1 room2 room3  - room
    hotel1 hotel2 - hotel
    single double - r_type
    oceanView gardenView - r_view ) 
    
  (:init 
    (available_room_hotel  room1 hotel1)
    (available_room_hotel  room2 hotel1)
    (available_room_hotel  room3 hotel2)
    (roomType room1  single)
    (roomType room2  double)
    (roomType room3  single)
    (roomView room1  oceanView)
    (roomView room2  gardenView)
    (roomView room3  oceanView)
    (difRoomType single double)
    (difRoomType double single) )
    
  (:goal 
   (booked_hotel single hotel1 oceanView) )   
)
\end{lstlisting}

\subsubsection{Abstraction of Alternative Sequences of Concrete Actions}  \label{subsubSection:absAltSeqActAppendix}

The low-level domain:

\noindent
\begin{lstlisting} [numbers=none, 
                    basicstyle=\scriptsize\sffamily, 
										xleftmargin=0.2em, 
										columns=fixed,       
                    keepspaces=true,
										breaklines=true,
										postbreak=\mbox{\textcolor{blue}{$\hookrightarrow$}\space\space},
										label={lst:Cat4EdLLExt}]
 (define (domain education01_LL)

    (:requirements :strips :typing)

    (:types
      workshop slideDeck department lectureHall webConferenceSoftware template facultyMember adjunctProfessor - object)

    (:predicates
      (workshopUnderReview ?w - workshop ?d - department)
      (pendingWorkShopRequest ?w - workshop ?d - department)
      (approvedWorkShop ?w - workshop ?d - department)
      (emptySlides ?sd - slideDeck)
      (slideDeckWritten ?t - template ?sd - slideDeck ?w - workshop)
      (installed ?t - template)
      (newTemplate ?t - template)
      (lecturerSelected ?w - workshop)
      (lectureHallScheduled ?w - workshop ?lh - lectureHall)
      (installedVideoConferencing ?w - workshop ?s - webConferenceSoftware)
      (lecturedOnCampus ?w - workshop ?lh - lectureHall)
      (lecturedOnline ?w - workshop ?s - webConferenceSoftware)
      (workshopOffered ?w - workshop)
      (feedbackCollected ?w - workshop)    
      (assignedFacultyMember ?fm - facultyMember ?w - workshop)
      (hiredAdjunctProfessor ?ap - adjunctProfessor ?w - workshop))   

   (:action reviewWorkShop
        :parameters (?w - workshop ?d - department)
        :precondition (pendingWorkshopRequest ?w ?d)
        :effect (and (workshopUnderReview ?w ?d)
                     (not (pendingWorkShopRequest ?w ?d))) ) 

   (:action approveWorkShop
        :parameters (?w - workshop ?d - department)
        :precondition (workshopUnderReview ?w ?d)
        :effect (and (approvedWorkShop ?w ?d)) ) 

   (:action writeSlides
        :parameters (?t - template ?sd - slideDeck ?w - workshop ?d - department)
        :precondition (and (approvedWorkshop ?w ?d)
                           (installed ?t)
                           (emptySlides ?sd))  
        :effect (and (slideDeckWritten ?t ?sd ?w)
                     (not (emptySlides ?sd))) ) 

   (:action installNewTemplate
        :parameters (?t - template ?w - workshop ?d - department)
        :precondition (and (newTemplate ?t)
                           (approvedWorkShop ?w ?d))
        :effect (and (installed ?t) 
                     (not (newTemplate ?t))) )    
   
   (:action assignFacultyMember
        :parameters (?t - template ?sd - slideDeck ?w - workshop  ?fm - facultyMember)
        :precondition (slideDeckWritten ?t ?sd ?w)
        :effect (and  (assignedFacultyMember ?fm ?w )
                      (lecturerSelected ?w)) )

   (:action hireAdjunctProfessor
        :parameters (?t - template ?sd - slideDeck ?w - workshop ?ap - adjunctProfessor)
        :precondition (slideDeckWritten ?t ?sd ?w)
        :effect (and  (hiredAdjunctProfessor ?ap ?w)
                      (lecturerSelected ?w)) )

   (:action scheduleLectureHall
        :parameters (?w - workshop ?lh - lectureHall)
        :precondition (lecturerSelected ?w)
        :effect (lectureHallScheduled ?w ?lh) )    
    
   (:action lectureOnCampus
        :parameters (?w - workshop ?lh - lectureHall)
        :precondition (lectureHallScheduled ?w ?lh)
        :effect (and (lecturedOnCampus ?w ?lh)
                     (workshopOffered ?w)) ) 

   (:action installVideoConferencing
        :parameters (?w - workshop ?s - webConferenceSoftware)
        :precondition (lecturerSelected ?w)                         
        :effect (installedVideoConferencing ?w ?s) )

   (:action lectureOnline
        :parameters (?w - workshop ?s - webConferenceSoftware)
        :precondition (installedVideoConferencing ?w ?s)
        :effect (and (lecturedOnline ?w ?s)
                (workshopOffered ?w)) ) 

   (:action collectFeedback
        :parameters (?w - workshop)
        :precondition (workshopOffered ?w)                         
        :effect (feedbackCollected ?w) )    
)
\end{lstlisting}%

The low-level problem instance:

\noindent
\begin{lstlisting} [numbers=none, 
                    basicstyle=\scriptsize\sffamily, 
										xleftmargin=0.2em, 
										columns=fixed,       
                    keepspaces=true,
										breaklines=true,
										postbreak=\mbox{\textcolor{blue}{$\hookrightarrow$}\space\space},
										label={lst:Cat4EdProbLLExt}]
 (define (problem education01Problem_LL)
 (:domain education01_LL)

 (:objects 
      genAI econMath - workshop
      slides1 slides2 - slideDeck 
      eecs econ - department
      l1 l2 - lectureHall
      zoom teams - webConferenceSoftware
      t1 t2 - template
      mary john - facultyMember 
      harry tom - adjunctProfessor)    
 (:init 
     (pendingWorkShopRequest genAI eecs)
     (installed t1)
     (newTemplate t2)
     (emptySlides slides1))			
 (:goal 
     (feedbackCollected genAi))   
)
\end{lstlisting}%

The purpose of abstraction is specified as:

\noindent
\begin{lstlisting} [numbers=none, 
                    basicstyle=\scriptsize\sffamily, 
										xleftmargin=0.2em, 
										columns=fixed,       
                    keepspaces=true,
										breaklines=true,
										postbreak=\mbox{\textcolor{blue}{$\hookrightarrow$}\space\space},
										label={lst:Cat4EdLLPurpose}]
provide an abstraction of the concrete domain such that the high-level domain abstracts over:
*a1* processing a workshop request, 
*a2* appointing a workshop lecturer,
*a3* delivering a workshop session, and,
*a4* any information about slide templates
\end{lstlisting}%

The high-level domain based on the above purpose of abstraction:

\noindent
\begin{lstlisting} [numbers=none, 
                    basicstyle=\scriptsize\sffamily, 
										xleftmargin=0.2em, 
										columns=fixed,       
                    keepspaces=true,
										breaklines=true,
										postbreak=\mbox{\textcolor{blue}{$\hookrightarrow$}\space\space},
										label={lst:Cat4EdHLExt}]
(define (domain education01_HL)

    (:requirements :strips :typing)

    (:types
     workshop slideDeck department teachingPlatform lecturer - object)

   (:predicates
    (pendingWorkShopRequest ?w - workshop ?d - department)
    (advancedWorkShopRequest ?w - workshop ?d - department)
    (emptySlides ?sd - slideDeck)
    (slideDeckWritten ?sd - slideDeck ?w - workshop)
    (teachingCompleted ?w - workshop ?tp - teachingPlatform)
    (workshopOffered ?w - workshop)
    (lecturerSelected ?w - workshop)
    (feedbackCollected ?w - workshop)
    (appointedLecturer ?l - lecturer ?w - workshop))
   
   (:action advanceWorkShopRequest
        :parameters (?w - workshop ?d - department)
        :precondition (pendingWorkShopRequest ?w ?d)
        :effect (and (advancedWorkshopRequest ?w ?d)
                     (not (pendingWorkShopRequest ?w ?d))) 
   ) 
  
   (:action writeSlides
        :parameters (?sd - slideDeck ?w - workshop ?d - department)
        :precondition (and (advancedWorkshopRequest ?w ?d) 
                           (emptySlides ?sd))                           
        :effect (and  (slideDeckWritten ?sd ?w)
                      (not (emptySlides ?sd))) ) 
   
   (:action appointLecturer
        :parameters (?sd - slideDeck ?w - workshop ?l - lecturer)
        :precondition (slideDeckWritten ?sd ?w)
        :effect (and (appointedLecturer ?l ?w)
                     (lecturerSelected ?w)) )
  
   (:action conductWorkShop
        :parameters (?w - workshop ?tp - teachingPlatform)
        :precondition (lecturerSelected ?w)
        :effect (and (teachingCompleted ?w ?tp)
                     (workshopOffered ?w)) )    
    
   (:action collectFeedback
        :parameters (?w - workshop)
        :precondition (workshopOffered ?w)                         
        :effect  (feedbackCollected ?w) )  
)
\end{lstlisting}%

The high-level problem instance:

\noindent
\begin{lstlisting} [numbers=none, 
                    basicstyle=\scriptsize\sffamily, 
										xleftmargin=0.2em, 
										columns=fixed,       
                    keepspaces=true,
										breaklines=true,
										postbreak=\mbox{\textcolor{blue}{$\hookrightarrow$}\space},
										label={lst:Cat4EdProbHLExt}]
 (define (problem education01Problem_HL)
 (:domain education01_HL)

 (:objects 
      genAI econMath - workshop
      slides1 slides2 - slideDeck 
      eecs econ - department
      l1 l2 zoom teams - teachingPlatform
      mary john harry tom - lecturer)    
 (:init 
     (pendingWorkShopRequest genAI eecs)
      (emptySlides slides1) )    
 (:goal 
   (feedbackCollected genAI))   
)
\end{lstlisting}%

\clearpage
\subsection{Prompts} \label{subSec:Prompts}

\subsubsection{Abstraction of Alternative Concrete Actions: Zero-Shot} \label{subSec:AbsAltMapPrompts}
Figures \ref{fig:cat2SystemPromptA-P1} and \ref{fig:cat2SystemPromptA-P2} illustrate the first and second parts of the system prompt for this category. The user prompt is visualized in Figure \ref{fig:cat2AUserPrompt}.

\begin{figure}[h]
\begin{tcolorbox}[myboxstyle={Abstraction of Alternative Concrete Actions:  Zero-Shot Learning - System Prompt Part 1}]  
You are an expert in PDDL (Planning Domain Definition Language). Your task is to create abstract (high-level) PDDL domain and problem
files based on a more concrete (low-level) PDDL domain, problem file, and a specified purpose of abstraction.
\\
\\
Think step by step.
\\
\\
The goal is to simplify the domain and problem file by reducing specific details while maintaining overall functionality,
following these guidelines:
\\
\\
\#\#** Step 1: Abstracting the PDDL Domain **\#\#

1.1. Consider the low-level domain and the purpose of abstraction. Identify the parameters, types, predicates, and actions that are:\\
   - **Important** for the abstraction purpose and should be retained in the high-level domain.\\
   - **Unimportant** for the abstraction purpose and should be removed or generalized.
\\
\\
1.2 Generate abstract types, predicates, and actions:
\\
   - Combine or generalize related elements (actions, predicates, types) wherever possible to reduce complexity.
     For example, two types 'hotel' and 'airbnb' can be combined into 'accommodation', predicates 'booked\_hotel' and 'booked\_airbnb'
     can be abstracted to 'booked\_accommodation', and actions 'book\_airbnb' and 'book\_hotel' can be abstracted to 'book\_accommodation',
     and parameters in such abstracted actions and predicates must be adjusted to abstracted types.

   - When abstracting actions, predicates, or parameters, ensure that the terminology of the
     abstracted domain reflect the core purpose or theme of the domain, e.g., in a travel-focused domain, actions like 'book\_hotel' and
     'book\_airbnb' should abstract to 'book\_accommodation,' retaining the notion of travel planning.
\\
\\
1.3. Follow these rules:
\\
   - **Rule 1**: Minimize the number of types, predicates, and actions in the high-level domain.\\
   - **Rule 2**: Abstract types, predicates, and actions realistically; use commonsense reasoning to ensure that
                 abstracted actions and predicates reflect practical, high-level planning tasks.\\
   - **Rule 3**: Use terminology that preserves the focus of the domain (e.g., travel planning)\\
   - **Rule 4**: Ensure logical consistency: e.g., effects should not contradict one another\\
   - **Rule 5**: Ensure that the abstract domain complies with the STRIPS fragment of PDDL, using only `:strips` and `:typing` in the requirements.
\end{tcolorbox}
\caption{Abstraction of Alternative Concrete Actions: Zero-Shot Learning, System Prompt Part 1}
\label{fig:cat2SystemPromptA-P1}
\end{figure}

\begin{figure}[h]
\begin{tcolorbox}[myboxstyle={Abstraction of Alternative Concrete Actions: Zero-Shot Learning - System Prompt Part 2}]
\#\#** Step 2: Abstracting the PDDL Problem File **\#\#

2.1. Using the abstract domain generated in \#\# Step 1 \#\#, adapt the low-level problem file:
\\
   - Retain the structure and objects from the low-level problem, adjusting types to match the high-level domain.
   - Only use predicates, actions, and types present in the abstract domain you generated in \#\# Step 1 \#\#.
\\
\\
2.2. Follow these rules:
\\
   - **Rule 6**: You must not create new objects nor rename existing objects in the problem file;
                 you must use ALL the objects in the \#\# Low-level PDDL Problem File \#\#, and simply adjust their types
                 to match the high-level domain. \\
   - **Rule 7**: Ensure that the goal of the problem remains logically consistent with the abstract domain's purpose.
	
\end{tcolorbox}
\caption{Abstraction of Alternative Concrete Actions: Zero-Shot Learning, System Prompt Part 2}
\vspace{2em}
\label{fig:cat2SystemPromptA-P2}
\end{figure}

\begin{figure}[h]
\begin{tcolorbox}[myboxstyle={Abstraction of Alternative Concrete Actions: Zero-Shot Learning - User Prompt (User Query)}]
%
\#\# Description \#\#

[To be filled with  natural language description of a low-level domain]
\\

\#\# Low-Level Domain \#\#

[To be filled with PDDL specification of the low-level domain]
\\

\#\# Purpose of Abstraction \#\#:

[To be filled with natural language description of the purpose of abstraction]
\\

\#\# Low-Level Problem Instance \#\#

[To be filled with PDDL specification of a low-level problem instance]
\\
\\

Solution:
\end{tcolorbox}
\caption{Abstraction of Alternative Concrete Actions: Zero-Shot Learning, User Prompt (User Query)}
\vspace{2em}
\label{fig:cat2AUserPrompt}
\end{figure}

%
\subsubsection{Abstraction of Alternative Concrete Actions: One-Shot} \label{subSec:AbsAltEapPrompts} 
Figure \ref{fig:cat2SystemPromptOneShot-P1} shows the system prompt. The user prompt is visualized in Figures \ref{fig:cat2BUserPrompt} and \ref{fig:cat2BUserPrompt2}. Figures \ref{fig:cat2BAssistantPrompt} and \ref{fig:cat2BAssistantPrompt2} represent the first and second parts of assistant prompts for this category respectively. 


\begin{figure}[h]
\begin{tcolorbox}[myboxstyle={Abstraction of Alternative Concrete Actions: One-Shot Learning - System Prompt}]
You are an expert in PDDL (Planning Domain Definition Language) and your task is to generate the abstract (high-level) PDDL domain
 and problem instance given a more concrete (low-level) PDDL domain, problem instance and a purpose of abstraction.

User will provide an example named \#\# Case 1 \#\#.
Provide the solution for \#\# Case 2 \#\#. The goal is to simplify the domain and problem instance by reducing specific details while maintaining overall functionality, following these guidelines:\\

Think step by step.\\

\#\#** Step 1: Abstracting the PDDL Domain **\#\#

1.1. Consider the low-level domain and the purpose of abstraction. Identify the parameters, types, predicates, and actions that are:\\
   - **Important** for the abstraction purpose and should be retained in the high-level domain.\\
   - **Unimportant** for the abstraction purpose and should be removed or generalized.

1.2 Generate abstract types, predicates, and actions:\\
   - Combine or generalize related elements (actions, predicates, types) wherever possible to reduce complexity;
     parameters in such abstracted actions and predicates must be adjusted to abstracted types. \\
   - When abstracting actions, predicates, or parameters, ensure that the terminology of the
     abstracted domain reflect the core purpose or theme of the domain (e.g., travel planning).

1.3. You must abide by the following \#\# Domain Abstraction Rules \#\#:
   - **Rule 1**: Minimize the number of types, predicates, and actions in the high-level domain.\\
   - **Rule 2**: Abstract types, predicates, and actions realistically; use commonsense reasoning to ensure that
                 abstracted actions and predicates reflect practical, high-level planning tasks.\\
   - **Rule 3**: Use terminology that preserves the focus of the domain (e.g., traveling) \\
   - **Rule 4**: Ensure logical consistency: e.g., effects should not contradict one another \\
   - **Rule 5**: Ensure that the abstract domain complies with the STRIPS fragment of PDDL, using only `:strips` and `:typing` in the requirements.\\

\#\#** Step 2: Abstracting the PDDL Problem Instance **\#\#

2.1. Using the abstract domain generated in \#\# Step 1 \#\#, adapt the low-level problem instance:\\
   - Retain the structure and objects from the low-level problem, adjusting types to match the high-level domain.\\
   - Only use predicates, actions, and types present in the abstract domain you generated in \#\# Step 1 \#\# for \#\# Case 2 \#\#.

2.2. You must abide by the following \#\# Problem Instance Abstraction Rules \#\# :\\
   - **Rule 6**: You must not create new objects nor rename existing objects in the problem instance;
                 you must use ALL the objects in the \#\# Low-level PDDL Problem Instance \#\# for \#\# Case 2 \#\#, and simply adjust their types to match the high-level domain.\\
   - **Rule 7**: Ensure that the goal of the problem remains logically consistent with the abstract domain's purpose.

\end{tcolorbox}
\caption{Abstraction of Alternative Concrete Actions: One-Shot Learning, System Prompt}
\vspace{2em}
\label{fig:cat2SystemPromptOneShot-P1}
\end{figure}


\begin{figure}[h]
\begin{tcolorbox}[myboxstyle={Abstraction of Alternative Concrete Actions: One-Shot Learning - User Prompt (Part of One-Shot Example) Part 1}]
\#\# Case 1 \#\#:\\
\#\# Description \#\#  \\
Our PDDL domain is based on travel arrangements.
At a more concrete level (low-level):  in order to go to the destination,
one may book a flight or train, if a seat is available, and book a hotel or airbnb, provided a room is available.\\

\#\# Low-Level Domain \#\#

(define (domain travelArrange01\_LL)\\
    (:requirements :strips :typing)

    (:types
     hotel airbnb room flight trainRide seat - object)

    (:predicates\\
    (booked\_hotel  ?r -room  ?h - hotel)\\
    (booked\_airbnb  ?r - room ?ab - airbnb)\\
    (available\_room\_hotel  ?r  - room ?h - hotel)\\
    (available\_room\_airbnb  ?r - room ?ab - airbnb) \\
    (bookedHotelOrAirbnb) \\
    (available\_seat\_flight ?s - seat ?f - flight) \\
    (available\_seat\_trainRide ?s -seat ?t - trainRide) \\
    (booked\_flight ?s - seat ?f - flight)\\
    (booked\_trainRide ?s - seat ?t - trainRide)\\
    (bookedFlightOrTrainRide))

    (:action book\_hotel\\
        :parameters (?h - hotel ?r - room)\\
        :precondition (available\_room\_hotel ?r ?h)\\
        :effect (and (booked\_hotel ?r ?h)\\
                     (not (available\_room\_hotel ?r ?h))\\
                     (bookedHotelOrAirbnb)))

    (:action book\_airbnb\\
        :parameters (?ab - airbnb ?r - room)\\
        :precondition (available\_room\_airbnb ?r ?ab)\\
        :effect (and (booked\_airbnb ?r ?ab)\\
                     (not (available\_room\_airbnb ?r ?ab))\\
                     (bookedHotelOrAirbnb))) 

    (:action book\_flight\\
        :parameters (?f - flight ?s - seat)\\
        :precondition (available\_seat\_flight ?s ?f)\\
        :effect (and (booked\_flight ?s ?f)\\
                     (not (available\_seat\_flight ?s ?f))\\
                     (bookedFlightOrTrainRide)))

   (:action book\_trainRide\\
        :parameters (?t - trainRide ?s - seat)\\
        :precondition (available\_seat\_trainRide ?s ?t)\\
        :effect (and (booked\_trainRide ?s ?t)\\
                     (not (available\_seat\_trainRide ?s ?t))\\
                     (bookedFlightOrTrainRide)))
)
\end{tcolorbox}
\caption{Abstraction of Alternative Concrete Actions: One-Shot Learning, User Prompt (Part of One-Shot Example) Part 1}
\label{fig:cat2BUserPrompt}
\end{figure}

\begin{figure}[h]
\begin{tcolorbox}[myboxstyle={Abstraction of Alternative Concrete Actions: One-Shot Learning - User Prompt (Part of One-Shot Example) Part 2}]
\#\# Purpose of Abstraction \#\#:

provide an abstraction of the concrete domain such that the high-level domain abstracts over
booking various kinds of transportation and accommodation.\\

\#\# Low-Level Problem Instance \#\#

(define (problem travelArrange01Problem1\_LL)\\
(:domain travelArrange01\_LL)

(:objects\\
    room1 room2 room3 room4 - room\\
    hotel1 - hotel\\
    airbnb1 - airbnb\\
    seat1 seat2 seat3 seat4 - seat\\
    flight1 - flight\\
    trainRide1 - trainRide)

  (:init\\
    (available\_room\_hotel  room1 hotel1)\\
    (available\_room\_hotel  room2 hotel1)\\
    (available\_room\_airbnb  room3 airbnb1)\\
    (available\_room\_airbnb  room4 airbnb1)\\
    (available\_seat\_flight  seat1 flight1)\\
    (available\_seat\_flight  seat2 flight1)\\
    (available\_seat\_trainRide  seat3 trainRide1)\\
    (available\_seat\_trainRide  seat4 trainRide1))

  (:goal\\
   (and (bookedFlightOrTrainRide)
        (bookedHotelOrAirbnb))))\\

Solution:
\end{tcolorbox}
\caption{Abstraction of Alternative Concrete Actions: One-Shot Learning, User Prompt (Part of One-Shot Example) Part 2}
\label{fig:cat2BUserPrompt2}
\end{figure}


\begin{figure}[h]
\begin{tcolorbox}[myboxstyle={Abstraction of Alternative Concrete Actions: One-Shot Learning - Assistant Prompt (Part of One-Shot Example) Part 1}]
\# Generating the Abstract (High-Level) Domain \# for \#\# Case 1 \#\# 

Rationale: \\
1- Consider the purpose of abstraction: high-level domain abstracts over booking various kinds of transportation
and accommodation. We must follow instructions in **\#\# Step 1: Abstracting the PDDL Domain \#\#**.

2- Hotel and Airbnb are kinds of accommodation. Create a more general (abstract) PDDL type named 'accommodation' that will replace types hotel and airbnb. Similarly, create a more general (abstract) type named 'transportation' to replace types trainRide and flight. Other types (room and seat) that are not abstracted must remain in the high-level domain.

3- Create abstract action book\_accommodation which abstracts over alternative choice of book\_hotel and book\_airbnb.
Parameter ?r - room remains the same, but parameters which refer to types hotel or airbnb must be replaced with type accommodation.
Similarly, create abstract action book\_transportation which abstracts over alternative choice of book\_flight and book\_trainRide.
Keep parameter ?s - seat, but parameters which refer to types flight or trainRide must be replaced with type transportation.

4- Create abstract predicate booked\_accommodation which abstracts over alternative choice of booked\_hotel and booked\_airbnb,
and choose their parameters following the approach in point 3 above.
Similarly, create abstract predicate booked\_transportation, available\_seat and available\_room, and choose their
parameters as advised above.

5- Predicate doneBookingHotelOrAirbnb is used to facilitate disjunctive goal of booking either a hotel or
airbnb in the low-level. Similarly, doneBookingFlightOrTrainRide facilitates disjuntive goal of booking either a flight or trainRide.
Replace them by more abstract predicates doneBookingAccommodation and doneBookingTransportation respectively.\\

\#\# High Level Domain \#\# \\

(define (domain travelArrange01\_HL)\\
    (:requirements :strips :typing)

    (:types
     accommodation room transportation seat - object)

    (:predicates \\
      (booked\_accommodation  ?r -room  ?a - accommodation) \\
      (available\_room  ?r - room ?a - accommodation) \\
      (doneBookingAccommodation)\\
      (available\_seat ?s - seat ?tp - transportation)\\
      (booked\_transportation ?s - seat ?tp - transportation)\\
      (doneBookingTransportation))

    (:action book\_accommodation\\
        :parameters (?a - accommodation ?r - room)\\
        :precondition (available\_room ?r ?a) \\
        :effect (and (booked\_accommodation ?r ?a) \\
                     (not (available\_room ?r ?a)) \\
                     (doneBookingAccommodation) )) 
)
\end{tcolorbox}
\caption{Abstraction of Alternative Concrete Actions: One-Shot Learning - Assistant Prompt (Part of One-Shot Example) Part 1}
\label{fig:cat2BAssistantPrompt}
\end{figure}

\begin{figure}[h]
\begin{tcolorbox}[myboxstyle={Abstraction of Alternative Concrete Actions: One-Shot Learning - Assistant Prompt (Part of One-Shot Example) Part 2}]

    (:action book\_transportation\\
        :parameters (?tp - transportation ?s - seat)\\
        :precondition (available\_seat ?s ?tp)\\
        :effect (and (booked\_transportation ?s ?tp)\\
                     (not (available\_seat ?s ?tp))\\
                     (doneBookingTransportation)))\\

\# Generating the Abstract (High-Level) Problem Instance \# for \#\# Case 1 \#\#

Rationale: \\
We must use the generated high-level domain 'travelArrange01\_HL' and also
follow instructions in **\#\# Step 2: Abstracting the PDDL Problem Instance \#\#**.
All objects must remain the same and be included. Only predicates and types used are those existing in high-level
domain.\\

(define (problem travelArrange01Problem1\_HL)\\
(:domain travelArrange01\_HL)

 (:objects\\
    room1 room2 room3 room4 - room\\
    hotel1 airbnb1 - accommodation\\
    seat1 seat2 seat3 seat4 - seat\\
    flight1 trainRide1 - transportation)

  (:init
    (available\_room  room1 hotel1)\\
    (available\_room  room2 hotel1)\\
    (available\_room  room3 airbnb1)\\
    (available\_room  room4 airbnb1)\\
    (available\_seat  seat1 flight1)\\
    (available\_seat  seat2 flight1)\\
    (available\_seat  seat3 trainRide1)\\
    (available\_seat  seat4 trainRide1))

  (:goal\\
   (and (doneBookingAccommodation)\\
        (doneBookingTransportation)))
)
\end{tcolorbox}
\caption{Abstraction of Alternative Concrete Actions: One-Shot Learning, Assistant Prompt (Part of One-Shot Example) Part 2}
\label{fig:cat2BAssistantPrompt2}
\end{figure}

\subsubsection{Abstraction of Sequences of Concrete Actions} \label{subSec:AbsSeqPrompts}
Figures \ref{fig:cat3SystemPrompt-P1} and \ref{fig:cat3SystemPrompt-P2} show the first and second parts of the system prompt respectively. 

\begin{figure}[h]
\begin{tcolorbox}[myboxstyle={Abstraction of Sequences of Concrete Actions - System Prompt Part 1}]
You are an expert in PDDL (Planning Domain Definition Language) and your task is to generate the abstract (high-level) PDDL domain
 and problem instance given a more concrete (low-level) PDDL domain, problem instance and a purpose of abstraction.\\

User will provide an example named \#\# Case 1 \#\#. 
Provide the solution for \#\# Case 2 \#\#, which includes the high-level Domain, Problem Instance and the Rationale.\\

The goal is to simplify the domain and problem instance by reducing specific details while maintaining
overall functionality, following these guidelines:
\\
\\
Think step by step.
\\
\\
\#\#Step 1: Abstracting the PDDL Domain\#\#
\\
1.1. Analyze the low-level domain and the purpose of abstraction. You must consider ALL actions and predicates, and
     identify predicates and actions that must be adjusted according to the following criteria:\\
     **Abstracted**: Select predicates and actions that include details irrelevant to the abstraction purpose.
                     Merge these elements to generalize the domain while maintaining functional integrity.\\
     **Unchanged**: Retain predicates and actions essential to preserving the domain's core functionality and ensuring alignment 
                    with the abstraction's purpose.\\

1.2 Generate abstract predicates and actions:\\
   - Review the low-level PDDL domain and purpose of abstraction to identify important sequences of actions that should be merged into
     higher-level actions that achieve the same effect but eliminate unnecessary intermediate steps.\\
   - The abstraction should aggregate low-level actions but retain logical succession based on the steps (i.e. actions should follow
     the necessary order of execution, and abstracted actions must enforce these logical constraints).\\
   - Create high-level predicates by combining (conjunctions of) low-level predicates when possible
     and adjust the parameter lists accordingly. \\

1.3. You must abide by the following \#\# Domain Abstraction Rules \#\#:\\
   - **Rule 1**: Minimize the number of predicates and actions in the high-level domain. Do not create new types;
                 retain all types from the low-level domain.\\
   - **Rule 2**: Abstract predicates and actions realistically; use commonsense reasoning to ensure that
                 abstracted actions and predicates reflect practical, high-level planning tasks.\\
   - **Rule 3**: When creating abstract actions and predicates, use terminology that preserves the focus of the domain
                 (e.g., online word editor).\\
   - **Rule 4**: Ensure logical consistency: e.g., effects should not contradict one another.\\
   - **Rule 5**: Ensure that the abstract domain complies with the STRIPS fragment of PDDL, using only `:strips` and `:typing` in the requirements. E.g., preconditions in STRIPS must not include negation (not)
\end{tcolorbox}
\caption{Abstraction of Sequences of Concrete Actions - System Prompt Part 1}
\label{fig:cat3SystemPrompt-P1}
\end{figure}

\begin{figure}[h]
\begin{tcolorbox}[myboxstyle={Abstraction of Sequences of Concrete Actions - System Prompt Part 2}]
\#\#Step 2: Abstracting the PDDL Problem Instance\#\#
\\
\\
2.1. Using the abstract domain generated for \#\# Case 2 \#\# in \#\# Step 1 \#\#, adapt the low-level problem instance:

   - Retain the structure and objects from the low-level problem of \#\# Case 2 \#\#. \\
   - Only use predicates, actions, and types present in the abstract domain you generated in \#\# Step 1 \#\#.

2.2. Follow these rules:\\
   - **Rule 6**: You must not create new objects nor rename existing objects in the problem instance;
                 you must use ALL the objects in the \#\# Low-level PDDL Problem Instance \#\#.\\
   - **Rule 7**: Account for all possible combinations of related objects when newly abstracted predicates are created in the high-level domain, especially when multiple types of objects are involved.\\
   - **Rule 8**: Ensure that the goal of the problem remains logically consistent with the abstract domain's purpose.

\end{tcolorbox}
\caption{Abstraction of Sequences of Concrete Actions - System Prompt Part 2}
\label{fig:cat3SystemPrompt-P2}
\end{figure}

\subsubsection{Abstraction of Action/Predicate Parameters} \label{subSec:AbsFeaturesPrompts}
Figure \ref{fig:cat1SystemPrompt} visualizes the system prompt, and the user prompt is shown in Figure \ref{fig:cat1UserPrompt}. 

\noindent
\begin{figure}[h]
\begin{tcolorbox}[myboxstyle={Abstraction of Action/Predicate Parameters - System Prompt}]
You are an expert in PDDL (Planning Domain Definition Language) and your task is to write an abstract (high-level) PDDL domain and problem file, given a 1) more concrete (low-level) domain and problem file and 2) a purpose for abstraction.
\\ 
\\
Let's think step by step.
\\
\\
First consider the purpose of abstraction and distinguish between 1) what details (represented by parameters, types, predicates, actions) are important wrt this purpose and must remain in the high-level domain and problem file and 2) which details (represented by parameters, types, predicates, actions) are not important and must be removed from the high-level domain and problem file.
\\
\\
You must not create any new actions, types, predicates, parameters nor objects. You must not rename any existing PDDL objects in the problem file.
\end{tcolorbox}
\caption{Abstraction of Action/Predicate Parameters - System Prompt}
\label{fig:cat1SystemPrompt}
\end{figure}


\begin{figure}[h]
\begin{tcolorbox}[myboxstyle={Abstraction of Action/Predicate Parameters - User Prompt (User Query)}]
\#\# PDDL Domain \#\#: is based on hotel bookings.
At a more concrete level (low-level): A hotel room has a room type (eg. single or double) and a room view (like ocean view or garden view).
It is possible to change a room type from one type to another (eg. from single to double)\\

(define (domain travelArrange02\_LL)\\
    (:requirements :strips :typing)

    (:types hotel r\_view room r\_type - object)

    (:predicates\\
    (booked\_hotel   ?rt - r\_type ?h - hotel ?rv - r\_view)\\
    (available\_room\_hotel  ?r  - room ?h -hotel)\\
    (roomType ?r - room ?rt - r\_type) \\
    (roomView ?r - room ?rv - r\_view)\\
    (difRoomType ?rt1 - r\_type ?rt2 - r\_type))

    (:action book\_hotel\\
        :parameters (?h - hotel ?r - room ?rv - r\_view ?rt - r\_type)\\
        :precondition (and (available\_room\_hotel ?r ?h)\\
                           (roomType ?r ?rt)\\
                           (roomView ?r ?rv) )\\
       :effect (and (booked\_hotel ?rt ?h ?rv)\\
                     (not (available\_room\_hotel ?r ?h)))
    )

    (:action change\_RoomType\\
        :parameters (?r - room ?rt1 ?rt2 - r\_type)\\
        :precondition (and(roomType ?r ?rt1)\\
                          (difRoomType ?rt1 ?rt2))\\
        :effect (and (roomType ?r ?rt2)\\
                     (not (roomType ?r ?rt1)))
        )
)


\#\# PDDL Problem \#\#:\\

(define (problem travelArrange02Problem1\_LL)\\
(:domain travelArrange02\_LL)

(:objects\\
    room1 room2 room3  - room\\
    hotel1 hotel2 - hotel\\
    single double - r\_type\\
    oceanView gardenView - r\_view)

  (:init\\
    (available\_room\_hotel  room1 hotel1)\\
    (available\_room\_hotel  room2 hotel1)\\
    (available\_room\_hotel  room3 hotel2)\\
    (roomType room1  single)\\
    (roomType room2  double)\\
    (roomType room3  single)\\
    (roomView room1  oceanView)\\
    (roomView room2  gardenView)\\
    (roomView room3  oceanView)\\
    (difRoomType single double)\\
    (difRoomType double single))

  (:goal\\
   (booked\_hotel single hotel1 oceanView))
)\\

\#\# Purpose of Abstraction \#\#:
provide an abstraction of the concrete domain such that the high-level domain must not include information about room view.
\end{tcolorbox}
\caption{Abstraction of Action/Predicate Parameters - User Prompt (User Query)}
\label{fig:cat1UserPrompt}
\end{figure}



\subsubsection{Abstraction of Alternative Sequences of Concrete Actions} \label{subSec:AbsALtSeqPrompts}
System prompt is illustrated in Figures \ref{fig:cat4SystemPrompt1} and \ref{fig:cat4SystemPrompt2}.

\noindent
\begin{figure}[h]
\begin{tcolorbox}[myboxstyle={Abstraction of Alternative Sequences of Concrete Actions - System Prompt Part 1}]
You are an expert in PDDL (Planning Domain Definition Language). Your task is to create abstract (high-level) PDDL domain and problem
files based on a more concrete (low-level) PDDL domain, problem file, and a specified purpose of abstraction.

User will provide an example named \#\# Case 1 \#\#. Provide the solution for \#\# Case 2 \#\#, which includes the high-level Domain, Problem Instance and the Rationale.\\

Think step by step.\\

The goal is to simplify the domain and problem instance by reducing specific details while maintaining
overall functionality, following below guidelines:\\

\#\#** Step 1: Abstracting the PDDL Domain **\#\# \\
1.1 Analyze the purpose of abstraction, and in the low-level domain, identify types, predicates and actions that must be adjusted according to the below criteria, in the order from 1.1.1 to 1.1.3:

     1.1.1 **Removed**:    Identify types, predicates and actions that must be removed as the abstraction purpose requires to abstract over  
		                       any information regarding them.

     1.1.2 **Abstracted**: Identify types, predicates, and actions that are assumed too detailed wrt the abstraction purpose;
                           merge these elements as guided in Step 1.2 below.  
     
     1.1.3 **Unchanged**:  Retain types, predicates and actions which must not be removed or abstracted, as in 1.1.1 and 1.1.2 above. \\
                          
1.2 Consider the purpose of abstraction and generate **Abstracted** types, predicates, and actions, which abstract over *Concrete* types, predicates and actions respectively, as instructed below: 

    1.2.1  Identify *alternative* related types (resp. actions, predicates) and generalize the types (resp. actions, predicates) wherever possible to reduce  complexity. E.g., type 'facutlyMember' and 'adjunctProfessor' are abstracted to 'lecturer', or, e.g., in context of teaching, as  types 'webConferenceSoftware' and 'lectureHall' can both be used for teaching and hence abstracted to type 'teachingPlatform'. Actions and predicates can be abstracted similarly.  
 
    1.2.2  Identify *sequences of low-level actions* that should be merged into a higher-level action that achieve the same effect but eliminate unnecessary intermediate steps. The abstraction should aggregate low-level actions but retain logical succession based on the steps (i.e. actions  should follow the necessary order of execution, and abstracted actions must enforce these logical constraints). 
           Parameters of the abstract action include the aggregation of parameters in low-level actions. Get inspirations from  \#\# Case 1 \#\# on how to generate high-level predicates which may be used as preconditions (resp. effects) of the generated high-level action, and which abstract over logical conjunction of certain existing preconditions (resp. effects).

\end{tcolorbox}
\caption{Abstraction of Alternative Sequences of Concrete Actions - System Prompt Part 1}
\vspace{2em}
\label{fig:cat4SystemPrompt1}
\end{figure}

\noindent
\begin{figure}[h]
\begin{tcolorbox}[myboxstyle={Abstraction of Alternative Sequences of Concrete Actions - System Prompt Part 2}]

1.3. You MUST abide by the following \#\# Domain Abstraction Rules \#\#: \\
   - **Rule 1**: Minimize the number of types, predicates and actions in the high-level domain. \\
   - **Rule 2**: Use commonsense reasoning to ensure that abstracted types, actions and predicates reflect practical, high-level planning tasks.\\
   - **Rule 3**: When creating abstract types, actions, and predicates, use terminology that preserves the focus of the domain. \\                
   - **Rule 4**: Ensure logical consistency: e.g., effects should not contradict one another.\\
   - **Rule 5**: Ensure that the abstract domain complies with the STRIPS fragment of PDDL, using only '(:requirements :strips :typing)'.
                 E.g., preconditions in STRIPS must not include negation 'not', and use of keywords like 'when', 'or', 'either' are forbidden. \\  
   - **Rule 6**: When a high-level action is generated, it is important to adapt its precondition and effects (possibly by using abstract predicates). \\

\#\#** Step 2: Abstracting the PDDL Problem File **\#\# \\

2.1. Using the abstract domain generated for \#\# Case 2 \#\# in \#\# Step 1 \#\#, adapt the low-level problem instance: \\
   - Retain the structure from the low-level problem of \#\# Case 2 \#\#.\\
   - Only use types, predicates, and actions present in the abstract domain you generated in \#\# Step 1 \#\#. \\

2.2. Follow these rules: \\
   - **Rule 7**: You must not create new objects nor rename existing objects in the problem instance. If a type for an object has been abstracted, objects must be assigned to their abstract type. If a type for an object has been removed following 1.1.1, the object must be removed.\\
   - **Rule 8**: Account for all possible combinations of related objects when newly abstracted predicates are created in the high-level domain, especially when multiple types of objects are involved. \\
   - **Rule 9**: Ensure that the goal of the problem remains logically consistent with the abstract domain's purpose.
\end{tcolorbox}
\caption{Abstraction of Alternative Sequences of Concrete Actions - System Prompt Part 2}
\vspace{2em}
\label{fig:cat4SystemPrompt2}
\end{figure}


\end{document}